\documentclass[10pt,a4paper]{article}
\usepackage{latexsym}
\usepackage{lscape}
\usepackage{graphicx}
\usepackage{hyperref}
\usepackage{microtype}
\usepackage{eurosym}
\usepackage{authblk}

\newtheorem{problem}{Problem} 

\newtheorem{LEMMA}{Lemma}
\newenvironment{lemma}{\begin{LEMMA} \hspace{-.85em} {\bf} \rm}%
                            {\end{LEMMA}}
\newtheorem{EXAMPLE}{Example}
\newenvironment{example}{\begin{EXAMPLE} \hspace{-.85em} {\bf} \rm}%
                            {\end{EXAMPLE}}
          
\usepackage{bm}      
\usepackage{amsmath}   
\usepackage{enumitem}

\title{Playing with and against Hedge}
\author[1]{Miltiades E. Anagnostou}
\affil[1]{School of Electrical and Computer Engineering, National Technical University of Athens,
 Zografou 15780, Greece\\ email:\href{mailto:miltos@central.ntua.gr}{miltos@central.ntua.gr}}
\author[2]{Maria A. Lambrou}
\affil[2]{Dept. of Shipping Trade and Transport, University of the Aegean,
Chios 82132, Greece\\
email:\href{mailto:mlambrou@aegean.gr}{mlambrou@aegean.gr}
}

\begin{document}



\maketitle

\begin{abstract}
Hedge has been proposed as an adaptive scheme, 
which guides an agent's decision in resource selection and distribution problems 
that can be modeled as a multi-armed bandit full information game.
Such problems are encountered
in the areas of computer and communication networks, 
e.g. network path selection, load distribution, network interdiction, and also in problems in the area of transportation. 
We study Hedge under the assumption that the total loss that can be suffered by the player in each round is upper bounded. In this paper, we study the worst performance of Hedge.

\end{abstract}

\newpage
\tableofcontents
\newpage
\section{Introduction}
\label{sect:introduction}

\subsection{The game model}
Firstly we summarize the generalized full information multi-armed bandit game:
A machine offers $N$ betting options (or ``arms'').
In each round the player bets an amount of money $M$ by assigning a fraction $p_i$ ($\sum_{i=1}^N p_i =1$) of  $M$ to option $i$.
The machine responds by determining a penalty (or reward) factor $\ell_i$ for each option $i$; effectively the player's result in the current round is equal to 
\begin{equation}
R= \sum_{i=1}^N p_i M \times \ell_i = M \sum_{i=1}^N p_i  \times \ell_i 
\end{equation}
The game is played for a pre-determined total number of $T$ rounds.
The player aims at minimizing the total loss or maximizing the total reward over the whole game.

This game is known as the {\em full information} version of the {multi-armed bandit game} \cite{Robbins1985} since the full penalty vector $(\ell_1, \ell_2, \ldots, \ell_N)$ is announced to the player after each round, the other option being to announce only the outcome $R$. 
Also, the game is called {\em generalized} in the sense that the player is allowed to play mixed strategies, i.e. to distribute the total bet to different options in the same round.

An ordinary assumption is that the bet per round $M$ remains constant; in the following, we shall set $M=1$ without loss of generality. However, a less ordinary assumption that we use in this paper is that {\em the loss that can be inflicted on the player in each round is upper bounded}. We express this restriction by upper-bounding the sum of penalties. Again, if we normalize the penalties w.r.t. to the upper bound, we can write
\begin{equation}
\sum_{i=1}^N  \ell_i \leq 1
\label{eq:upperboundedloss} 
\end{equation}
In other words the machine works within a limited penalty budget in each round.
Effectively the normalization assumption $\sum_i p_i =1$ together with (\ref{eq:upperboundedloss}) imply that $\sum_{i=1}^N  \ell_i p_i \leq \max_i \{ \ell_i \} \leq 1$, thereby restricting the maximum loss per round to one unit. 
The $\ell_i$s can become rewards by assigning them negative and restricting their sum to be greater than~-1. In this paper we use losses.

\subsection{A fair game interpretation}
The loss based game seems unfair, since the player always suffers losses, but it can easily be converted to a fair game: Suppose that  in each round the player is awarded with an amount equal to $1/N$, therefore the gain of the player in a single round is 
\[
g=\frac{1}{N}-\sum_{i=1}^N p_i  \times \ell_i .
\]
Then the player could choose $p_1=p_2=\ldots=p_N=1/N$, which implies $g=0$. Effectively, if the player insists on bet equipartition, no positive gains can be achieved either. If the player can predict the option with the minimum loss that will appear in the next round, say option $i_0$, the gain will become $g=1/N- \ell_{i_0} > 0$ (since $\min_i \ell_i < \sum_i \ell_i =1/N$) by choosing $p_{i_0}=1$. For an arbitrary bet distribution vector $(p_1, \ldots, p_N)$ the gain $g$ can be either positive or negative. The worst possible gain outcome is $g = 1/N - \max_i \ell_i$, and will appear if the player is extremely unlucky, so as to bet exclusively on the worst option. 

It can also be shown that a uniformly random selection of weights $p_i$ and losses $\ell_i$ will produce an average loss equal to $1/N$, thus a neutral outcome of the gain. Any successful prediction can tilt the balance of the game in favor of the player. A well known fact in the literature  is that if the arms of the bandit behave according to stationary processes, the Hedge algorithm (see Section~\ref{sec:hadgeintro}) brings the per round gain asymptotically to the optimum, i.e. towards  $1/N- \min_i E\{ \ell_i \}$. Effectively, Hedge will take advantage of any difference in the statistical behavior of the arms, and due to \ref{eq:upperboundedloss} it will produce positive gains.

However, in this paper we explore the performance of Hedge when there is no stationarity or other assumption that restricts the multi-armed bandit behavior. In fact, we determine the behavior of the bandit that results into the worst possible outcome of the player, when the player has adopted Hedge as a decision making aid. 

\subsection{The player's hand and Hedge algorithm}
\label{sec:hadgeintro}
The player aims at minimizing the total loss over the whole game; 
this aim can only be achieved by suitably controlling the bet distribution $p_i$ in each round.
Clearly, if the player could predict the result of the current round, she would choose the option with the minimum loss and bet the available unit of money on this option.
However, the player can only rely on past outcomes. 

This is where the {\em Hedge} or {\em multiplicative updates} algorithm   \cite{schapire1995desicion} enters in order to possibly guide the player's hand.
Hedge maintains a vector $w^t=(w_1^t, w_2^t, \ldots, w_n^t)$ of weights, such that $w_i^t \geq 0$ ($t=0,1,\ldots, T-1$, and $i=1,2, \ldots, N$). 
 In each round $t$ Hedge chooses the bet allocation to be
\begin{equation}
p_i^t= \frac{w_i^t}{\sum_{i=1}^N w_i^t}
\end{equation}
When the $t$'th round loss vector is revealed, $w^t$ is updated by using
\begin{equation}
w_i^{t+1}= w_i^t   \beta^{\ell_i^t }
\end{equation}
for some fixed $\beta$, such that $0 \leq \beta \leq 1$.

Our objective is to find {\em the worst possible performance of Hedge} over all possible responses of the multi-armed bandit during a game of fixed duration. 

\subsection{Overview of previous results}
In \cite{auer1995gambling} Auer et al. have proved that the performance of Hedge is asymptotically optimal for the {\em non} generalized game, the error being  $O(\sqrt{T N \ln N})$.
Note that the theorem holds for players that do not use mixed strategies. 
In the ``optimal'' game the player is supposed to choose the same overall most favorable option in each round, i.e. the option with the lowest loss over the whole game. 

The model used in \cite{auer1995gambling} effectively considers a machine with $N$ arms that is somehow imperfect and produces statistically unbalanced results. In a  perfect machine all arms would be identical, but in an imperfect machine there is an arm that statistically produces the most favorable results. The intention of the player is to discover exactly this arm. Since the player cannot predict the outcome of any arm in a specific round, the optimal strategy of the player would be to always bet on the same most favorable arm. It would take possibly a lot of time and money before a real player would be able to gather enough statistics, that will determine the best arm. The aforementioned theorem guarantees that Hedge somehow digs out this information for the player that has no previous experience with the machine.

Hedge has become quite popular recently and in different fields. 
For example, Chastain et al.\ have shown that Hedge is also applicable in coordination games and evolution theory and have noted that this algorithm ``has surprised us time and again with its seemingly miraculous performance and applicability''~\cite{chastain2014algorithms}.

There is also a number of Hedge variations and alternatives. Auer, Cesa-Bianchi et al.\ have proposed the {\em Exp3} algorithm in \cite{auer2002nonstochastic}.
Allenberg-Neeman and Neeman proposed  the {\em GL} (Gain-Loss) algorithm, for the full information game  \cite{allenberg2004full}.
Dani, Hayes, and Kakade have proposed the {\sc \em GeometricHedge} algorithm in \cite{dani2008price}. A modification was also proposed by Bartlett, Dani et al.\ in \cite{bartlett2008high}.
Cesa-Bianchi and Lugosi have proposed the {\sc ComBand}  algorithm  for the bandit version \cite{cesa2012combinatorial}. A comparison of algorithms can be found in the paper by Uchiya et al.\ \cite{uchiya2010algorithms}.


In two previous works of ours \cite{anagnostou2014playing,anagnostou2015worst} we have mentioned a number of application areas and problems, that could make use of the multi-armed bandit game with upper-bounded loss in a each round, i.e. as modified by  (\ref{eq:upperboundedloss}). Possible applications include path selection in computer networking, and network interdiction with applications in the area of transportation and border control.

In the aforementioned works we have studied the worst possible performance of Hedge (when it guides the player's hand in the aforementioned multi-armed bandit game) by exploring the penalty sequence vectors that maximize the player's total loss. 
In particular, in \cite{anagnostou2014playing} we have given a methodology for the calculation of the maximum total loss that is suffered by the player, and also for the calculation of the associated penalty vectors that are produced by the multi-armed bandit machine. We have also provided a formula for the total loss and the associated penalties that is a recursive function of the number of rounds and can be exploited for the numerical calculation of the performance of multiple round games. By using this recursive calculation we can get a fair idea of the ``optimal'' behavior of the machine (which is of course worst for the player). 
In general the resulting optimal  penalty vectors are real valued, but it appears that they become binary after a number of rounds.
In \cite{anagnostou2014playing} we have shown that if the penalties are restricted to binary values, the machine's optimal plan (worst for the player) can be easily calculated by using a greedy algorithm. A consequence of greediness is that the penalty plan becomes periodic, and the losses suffered by the player become also  periodic.

In \cite{anagnostou2015worst} we have given details  and examples of analytic calculation of the optimal plan from the machine's viewpoint, together with the resulting total loss. The results have shown that the optimal penalties are generally non binary, at least in the first few rounds. However, the calculation of the optimal penalty plan in terms of closed form formulas appears to be partly infeasible and partly too complicated as the  number of rounds~$T$ of the game increases. Complexity shows on the shape of the maximum total loss, which is a continuous curve with lots of kinks, i.e. points of sudden slope change. 

\subsection{New contributions}
In summary, in our past work (a)  we have provided numerical results and examples, (b) we have shown that the optimal binary penalty scheme is greedy, and (c) we have provided accurate results for games of short duration (up to three rounds).

In this paper we have more or less completed the analysis of the aforementioned problem: 
\begin{enumerate}
\item We  show how a game of any number of rounds can be analyzed accurately, by reducing its solution to a  single variable optimization problem. 
\item We calculate the distance between the optimal solution and a greedy binary solution, and we show that the binary solution is actually a very good approximation.
\item We describe the general ``nature'' of the optimal solution, which is greedy if the number of rounds does not exceed a well defined limit, but past this limit the optimal scheme evolves in an optimized periodic scheme.
Effectively, the complexity of the problem is polynomial. 
\item We give an explicit solution if the player starts Hedge with equal arm weights.
\end{enumerate} 
In summary, although some minor details can still be filled in, we give a full solution to the problem of the worst performance of Hedge when the multi-armed bandit functions under a limited penalty budget.

Finally a note on the methodology used in this paper:
In order to calculate the worst performance of Hedge we try to determine the bandit machine behavior that is most unfavorable to the player. This approach is consistent with the so called ``adversarial analysis'' of online algorithms.
According to S. Irani and A. Karlin (in section 13.3.1 of  \cite{hochbaum1996approximation}) a technique in finding bounds is to use an ``adversary'' who plays against an algorithm $\mathcal{A}$ and concocts an input, which forces $\mathcal{A}$ to incur a high cost. 
In our analysis  the adversary  tries to maximize Hedge's total loss by controling the penalty vector.

Although an ``adversary'' is a fictional methodological aid for the aforementioned analysis, there are applications and problem interpretations that make use of a real adversary. In \cite{anagnostou2014playing} we have described  a collection of such applications, e.g. network interdiction and border control.

\section{The loss maximization problem}
\subsection{Problem formulation}

The ``adversary'' controls the bandit machine so as to maximize the player's losses. A naive adversary would probably use the greedy approach, i.e. the adversary would maximally penalize the option with the highest bet, i.e. in round $t$ the adversary would  set $\ell_j^t =1$ for the option $j$, such that $p_j^t = \max_{i=1,\ldots,N} p_i^t$. We shall see later that this approach is less naive than expected. 
However, in principle a more sophisticated adversary would be expected to solve an optimization problem that takes into account the whole duration of the game.
We assume that the ``adversary'' is aware of the fact that the player is playing Hedge and is also aware of the value of the adjustment speed parameter $\beta$. Then the adversary has to solve the following optimization problem \cite{anagnostou2014playing}:

\begin{problem} \label{problem:general1}
Given a number of options $N$, an initial normalized weight vector  $\boldsymbol{w}=(w_1 , w_2 , \ldots , w_N )$, and a Hedge parameter $\beta$,  find the sequence $\boldsymbol{\ell}^0$, $\boldsymbol{\ell}^1, \ldots$, $\boldsymbol{\ell}^{T-1} $ that maximizes 
the player's total cumulative loss
\begin{equation}
L_{H(\beta)} = \sum_{t=0}^{T-1} \sum_{i=1}^N p_i^t \ell_i^t
\label{eq:someoflosses}
\end{equation}
 where
$ \boldsymbol{\ell}^t = (\ell_1^t, \ldots , \ell_N^t) $ is the penalty vector in round $t$ ($t=0, 1, \ldots, T-1$), 
such that $\sum_{i=1}^N \ell_i^t = 1$, 
  and the penalty weights $p_i^t$ in round $t$ are updated according to Hedge, i.e.
\begin{equation} 
 w_i^t = w_i^{t-1} \beta^{\ell_i^{t-1}}
, \quad 
p_i^t = \frac{w_i^t}{\sum_{i=1}^N w_i^t}  \quad (t \geq 1)
 \label{eq:changeWeight}
\end{equation} 
 for $t=1,\ldots,T-1,$ and $p_i^0=w_i$.~$\Box$
\end{problem}
 Clearly  the objective function (\ref{eq:someoflosses})
 is a function of  the $N$ initial weights $w_i$, the $N \times T$ variables $\ell_i^t$, and $\beta$.  Due to the normalization of weights and penalties there are   in total $(N-1) \times (T+1) +1$  independent variables.
The player chooses the $N-1$ independent initial weights, and $\beta$, and lets Hedge make all future decisions.  The $(N-1) \times T$ independent penalty variables are under the control of the bandit machine, i.e. the adversary.
In the following we     use 
$ L^{T-1}(w_1,  \ldots, w_N; \ell_1^0,  \ldots, \ell_N^0, \ldots, \ell_1^{T-1}, \ldots, \ell_N^{T-1}) $
or $L^{T-1}(\boldsymbol{w};\boldsymbol{\ell^0},\ldots, \boldsymbol{\ell^{T-1}})
$ instead of $L_{H(\beta)} $
whenever it is necessary to refer to  these variables.

For the adversary problem~\ref{problem:general1} is a well defined maximization problem. Analysis should aim at finding the 
maximum value of $L_{H(\beta)}$, as given by (\ref{eq:someoflosses}), together with the penalty vectors $\boldsymbol{\ell}^0$, $\boldsymbol{\ell}^1, \ldots$, $\boldsymbol{\ell}^{T-1} $ that achieve the maximum.  We show that an iterative solution is possible, but it hardly produces any closed form solutions. We shall also see that numerical results for $L_{H(\beta)}$ as a function of the initial weights.
The analysis of short games, i.e. games with a small number of rounds $T$, as given in \cite{anagnostou2015worst}, has shown that the optimal strategy of the adversary soon ends up with extreme penalty values, which in this case become binary due to normalization. This means that the optimal penalties, from the adversary's viewpoint, may take values between zero and one in the first few rounds, but soon the adversary is obliged to place all the available penalty (i.e. one unit) into a single option in each round. On the other hand, we have shown in \cite{anagnostou2014playing} that
the optimal strategy, under the assumption of binary penalties, is greedy (and simple to calculate). 
A straightforward conclusion that follows from the greediness is that the optimal binary strategy is also periodic. 
This means that the adversary assigns a unitary penalty to a different option in each round in a  cyclic manner, the cycle consisting of $N$ rounds, where $N$ is the number of options (provided of course that the game duration $T$ is long enough to accommodate more than one cycles). 

\section{First results}
In this section we reiterate some of the so far known results. Firstly, we give an outline of the  recursive properties of the optimal solution to Problem~\ref{problem:general1}. Then we use these properties in producing an initial set of numerical results.
\subsection{Iterative properties of the optimal solution}
\label{sect:recursion}
If in the current round the player (guided by Hedge) bets according to a weight vector $\boldsymbol{v}=(v_1, \ldots, v_N)$, and the adversary generates penalties $\boldsymbol{\ell}=(\ell_1,\ldots,\ell_N)$, then in the next round the player will use weights  
$\boldsymbol{W}(\boldsymbol{v},\boldsymbol{\ell}) = (W_1(\boldsymbol{v},\boldsymbol{\ell}), \ldots, W_N(\boldsymbol{v},\boldsymbol{\ell})), $ where
\begin{equation}
W_i(\boldsymbol{v},\boldsymbol{\ell}) = \frac{v_i \beta^{\ell_i}}{\sum_{j=1}^{N} v_j \beta^{\ell_j} } \quad (i=1,2,\ldots, N)
\label{eq:WdefinedGen}
\end{equation}
The total   loss of a $T$ round game, which starts with weights $\boldsymbol{w}$,  can be written as the sum of the losses of a single round game, which starts with weights $\boldsymbol{w}$, and a $T-1$ round game, which starts with weights $\boldsymbol{W}(\boldsymbol{w},\boldsymbol{\ell})=(W_1(\boldsymbol{w},\boldsymbol{\ell}) \ldots, W_N(\boldsymbol{w},\boldsymbol{\ell}))$, i.e.
\begin{equation}
L^{T-1} (\boldsymbol{w}; \boldsymbol{\ell}^0, \boldsymbol{\ell}^1, \ldots, \boldsymbol{\ell}^{T-1}) =
L^0 (\boldsymbol{w};\boldsymbol{\ell}^0) 
+  L^{T-2} \left(\boldsymbol{W}(\boldsymbol{w},\boldsymbol{\ell}^0); \boldsymbol{\ell}^1, \ldots, \boldsymbol{\ell}^{T-1} \right)  \label{eq:TgamesTotalLossRecursionGen1}
\end{equation}
Assuming that the solution to Problem~\ref{problem:general1} is
\[
L^{T-1}_{\max} (\boldsymbol{w}) =  \max_{\boldsymbol{\ell^0}, \ldots, \boldsymbol{\ell^{T-1}}} L^{T-1}(\boldsymbol{w};\boldsymbol{\ell^0},\ldots, \boldsymbol{\ell^{T-1}})
\]
the following iterative formula for $L^{T-1}_{\max} (\boldsymbol{w})$ can be derived from (\ref{eq:TgamesTotalLossRecursionGen1}):
\begin{equation}
L^{T-1}_{\max} (\boldsymbol{w}) = \max_{\boldsymbol{\ell}} \left[ L^0(\boldsymbol{w};\boldsymbol{\ell}) + L_{\max}^{T-2} (\boldsymbol{W}(\boldsymbol{w},\boldsymbol{\ell})) \right]
\label{eq:TgamesTotalLossRecursionVect}
\end{equation}
where $\boldsymbol{\ell^0}=\boldsymbol{\ell}$ is the penalty vector chosen by the adversary in the initial round.

The associated optimal penalties can also be computed iteratively. Let $\boldsymbol{\lambda}^{{T-1};t} (\boldsymbol{w})  = (\lambda^{{T-1};t}_1  (\boldsymbol{w}), \ldots, \lambda^{{T-1};t}_N  (\boldsymbol{w}))$, where $\lambda^{{T-1};t}_i  (\boldsymbol{w})$ denotes the optimal penalty of the $i$'th option in the $t$'th round of a $T$ round game  (starting with weights $\boldsymbol{w}$).
Then, the optimal penalties are given by the following formulas:
\begin{eqnarray}
\boldsymbol{\lambda}^{T-1;0}  (\boldsymbol{w}) 
&=& 
\arg \max_{\boldsymbol{\ell}} \left[ L^0(\boldsymbol{w};\boldsymbol{\ell}) + L_{\max}^{T-2} (\boldsymbol{W}(\boldsymbol{w},\boldsymbol{\ell})) \right] 
\label{eq:optimalLastPenaltyVect} \\
\boldsymbol{\lambda}^{T-1;t+1}  (\boldsymbol{w}) 
&=& 
\boldsymbol{\lambda}^{T-2;t} (\boldsymbol{W}(\boldsymbol{w},\boldsymbol{\lambda}^{T-1;0}(\boldsymbol{w}))) \quad (t=0, 1, \ldots, T-2)
\label{eq:optimalOtherPenaltiesVect} 
\end{eqnarray}
More details of the derivation together with sample calculations can be found in~\cite{anagnostou2014playing}. 
\subsection{Numerical results based on the iterative solution}
\begin{figure}[htb]
	\centering
  \includegraphics[width=.7\textwidth]{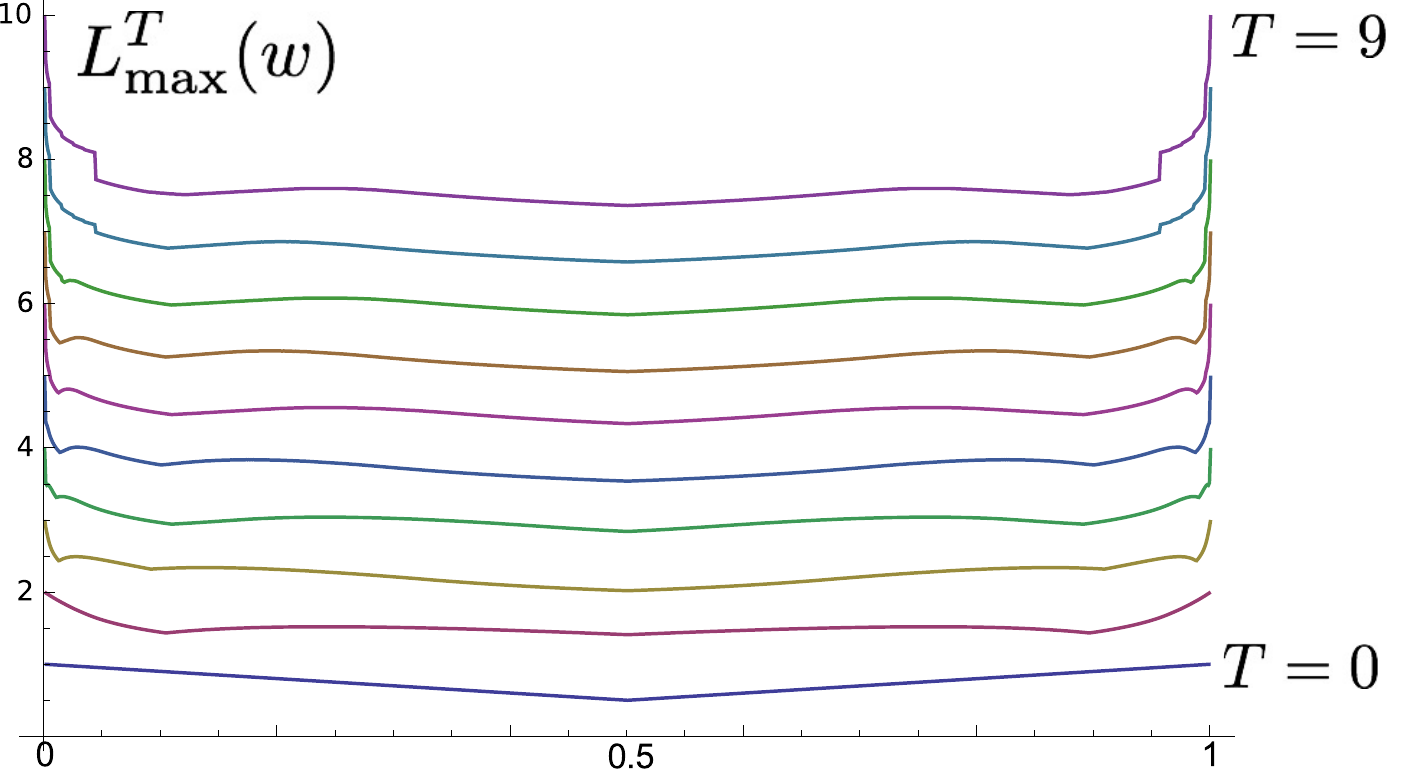}
  \caption{Plot of  $L^T_{\max}(w)$ (maximum loss of a $T+1$ round game) vs. $w$ for  $\beta=0.1$ and $T=0,1,\ldots ,9$.}
  \label{fig:QuantizedNumericalCalculationAcc1000B01V81gr11}
\end{figure}

The numerical calculation is based on sampling the initial weight vector $\boldsymbol{w}$. For example, if $N=2$, then the initial weights are, say, $w$ and $1-w$, and $L^{T-1}_{\max}(\boldsymbol{w}) = L^{T-1}_{\max}(w)$. The calculation starts with $L^{0}_{\max}(w)$ for samples of $w = k \Delta w$, and continues for $L^{1}_{\max}(w), L^{2}_{\max}(w), \ldots$ The accuracy of the outcome is conditional on the sampling resolution, i.e. on $\Delta w$, and a sufficiently accurate computation for a game with more rounds requires an increased resolution. We give an example of the outcome of such a computation in Fig.~\ref{fig:QuantizedNumericalCalculationAcc1000B01V81gr11}, 
in which we have used a rather exaggerated value of $\beta$ in order to expose the shapes of the curves more clearly. The curves are non linear and exhibit lots of sharp points. An example of the calculation of the associated penalties is given in Section~\ref{sec:optimalpenalties}.

Note that formula (\ref{eq:TgamesTotalLossRecursionVect})
does {\em not} imply that an optimal adversarial strategy for a $T-1$ round game is the same as the strategy of the first $T-1$ rounds in a $T$ round game, due to the fact that given an initial vector of weights, $L^{T-2}_{\max}$ enters the computation for $L^{T-1}_{\max}$ with a different vector of weights than the initial vector. In other words, the total number of rounds affects
the choice of penalties in the initial rounds too.

\subsection{Initial observations on the expected solution}

The obvious attack to the problem is to directly use(\ref{eq:TgamesTotalLossRecursionVect}) (and (\ref{eq:optimalOtherPenaltiesVect}) for the penalties).
In \cite{anagnostou2015worst} we have provided detailed results for $N=2$ option games with 1, 2, and 3 rounds ($T=0,1,2$). The results clearly illustrate the complexity of the analytic approach, mainly due to the fact that there is a population of sharp points of the function $L^T_{\max}(w)$ that grows with the number of rounds, and requires  
very long formulas for the calculation of the values of $w$ that produce the sharp points. Also, these results can possibly give some initial indications on the nature of the expected solution to the general problem.

\begin{example} \label{exmpl:equalweights} {\em Two and three round games starting with equal weights:}
Consider the case $w=1/2$, which means that the player begins the first round by equally distributing the bet to the two available options. Note that the choice $w=1/2$ is optimal for the player, as indicated by the results in \cite{anagnostou2015worst} and also by the numerical results of Fig.~\ref{fig:QuantizedNumericalCalculationAcc1000B01V81gr11}.
In a single round game the loss outcome for the player is $\frac{1}{2} \times \ell + \frac{1}{2} \times (1 - \ell) = \ell /2 $, i.e. constant regardless of the adversary's choice of penalties $\ell$ and $1- \ell$.
In a two round game the analysis in \cite{anagnostou2015worst} shows that the adversary should use penalties $\boldsymbol{\ell^0}=(0,1)$ in the first round and $\boldsymbol{\ell^1}=(1,0)$ in the second round (i.e. to put all the available penalty to the first option in the first round, and to the second option in the second round), or vice versa. By using this strategy the adversary will inflict on the player a total loss equal to 
$1/2 + 1/(1+ \beta)$.
So far the optimal penalties are binary. 

However, in a three round game the optimal policy for the adversary is
$
(\boldsymbol{\ell^0}, \boldsymbol{\ell^1}, \boldsymbol{\ell^2})
 = \left( (\frac{3}{4},\frac{1}{4}), (0,1), (1,0) \right)
$
or vice versa, i.e. $\left( (\frac{1}{4},\frac{3}{4}), (1,0), (0,1) \right)$. Both choices give a total loss equal to 
$1/2 + 2/(1 + \sqrt{\beta})$.
This time the first round penalties are fractional, while the 2nd and 3rd round penalties remain binary as before. This example shows  (a)~that the optimal penalties can be fractional and (b)~that the optimal penalties of any round depend on the total number of rounds.~$\Box$
\end{example}

\begin{example} {\em Periodic extension of the previous game:}
At this point let us make a few observations on the possible extension of the aforementioned two and three round schemes, which can serve as an introduction to periodic schemes. In Table~\ref{table1:extension} we show the loss that results from the periodic extension of the two round scheme, i.e. from the cyclic repetition of the penalties chosen by the adversary in the first two rounds. We present the next two rounds only, but the scheme could be cyclically extended to an arbitrary number of rounds. If this game is repeated for an even number of rounds, the average loss per round is equal to
\[
\frac{1}{2} \left( \frac{1}{2} + \frac{1}{1+ \beta} \right)
\]
Note that the choice of penalties could also be the outcome of a greedy algorithm for the adversary. Next, in Table~\ref{table2:extension}, we give the periodic extension of the second and third round of the three round example, i.e. we exclude the very first round from the periodic scheme. Note, again, that this could be a greedy scheme for the adversary. The loss per round in this second example (excluding the very first round)  is
\[
\frac{1}{1+ \sqrt{\beta}}
\]
and it is easy to prove that $\frac{1}{1+ \sqrt{\beta}} > \frac{1}{2} \left( \frac{1}{2} + \frac{1}{1+ \beta} \right)$ for $0< \beta < 1$. Thus, the cyclic strategy of Table~\ref{table2:extension} is superior to the cyclic strategy of Table~\ref{table1:extension} (from the adversary's viewpoint). Effectively, by resorting to a fractional penalty in the first round, the adversary is able to improve the long term behavior.~$\Box$
\end{example}

\begin{table}
\centering
\begin{tabular}{|c|c|c|c|c|c|}
\hline
round $t$ & $w_1^t$ & $w_2^t$ & $\ell_1^t$ & $\ell_2^t$ & loss in round $t$ \\ \hline
0 & $1/2$ & 1/2 & 1 & 0 & 1/2  \\
1 & $\frac{\frac{1}{2} \beta^1}{\frac{1}{2} \beta^1 + \frac{1}{2} \beta^0} = \frac{ \beta }{ 1 + \beta}$ &  
$\frac{\frac{1}{2} \beta^0}{\frac{1}{2} \beta^1 + \frac{1}{2} \beta^0} = \frac{ 1}{ 1 + \beta}$ & 0 & 1 & $1 / (1 + \beta)$ \\
2 & $\frac{ \frac{ \beta }{ 1 + \beta} \beta^0 }{\frac{ \beta }{ 1 + \beta} \beta^0 + \frac{ \beta }{ 1 + \beta} \beta^1} = \frac{1}{2}$ 
& $\frac{ \frac{ 1 }{ 1 + \beta} \beta^1 }{\frac{ \beta }{ 1 + \beta} \beta^0 + \frac{ 1}{ 1 + \beta} \beta^1} = \frac{1}{2}$ & 1 & 0 & 1/2  \\
3 & $ \beta / (1 + \beta)$ & $1 / (1 + \beta)$ & 0 & 1 & $1 / (1 + \beta)$ \\ \hline
\end{tabular}
\caption{Extension of the two round game} \label{table1:extension}
\end{table}

\begin{table}
\centering
\begin{tabular}{|c|c|c|c|c|c|}
\hline
round $t$ & $w_1^t$ & $w_2^t$ & $\ell_1^t$ & $\ell_2^t$ & loss  \\ \hline
0 & 1/2 & 1/2 & 1/4 & 3/4 & 1/2  \\
1 & $\frac{\frac{1}{2} \beta^{\frac{1}{4}}}{\frac{1}{2} \beta^{\frac{1}{4}}+ \frac{1}{2} \beta^{\frac{3}{4}}} = \frac{1}{1+ \sqrt{\beta}}$ 
& $\frac{\frac{1}{2} \beta^{\frac{3}{4}}}{\frac{1}{2} \beta^{\frac{1}{4}}+ \frac{1}{2} \beta^{\frac{3}{4}}} = \frac{\sqrt{\beta}}{1+ \sqrt{\beta}}$ 
& 1 & 0 & $\frac{1}{1+ \sqrt{\beta}}$ \\
2 & $\frac{\frac{1}{1+ \sqrt{\beta}} \beta^1}{\frac{1}{1+ \sqrt{\beta}} \beta^1 + \frac{\sqrt{\beta}}{1+ \sqrt{\beta}} \beta^0} = \frac{\sqrt{\beta}}{1+ \sqrt{\beta}}$ 
&  $\frac{\frac{\sqrt{\beta}}{1+ \sqrt{\beta}} \beta^0}{\frac{1}{1+ \sqrt{\beta}} \beta^1 + \frac{\sqrt{\beta}}{1+ \sqrt{\beta}} \beta^0} = \frac{1}{1+ \sqrt{\beta}}$
& 0 & 1 & $\frac{1}{1+ \sqrt{\beta}}$  \\
3 & $ \frac{1}{1+ \sqrt{\beta}}$ & $\frac{\sqrt{\beta}}{1+ \sqrt{\beta}}$ & 1 & 0 & $\frac{1}{1+ \sqrt{\beta}}$ \\ \hline
\end{tabular}
\caption{Extension of the three round game} \label{table2:extension}
\end{table}

\subsection{The binary case}
Analytic results from short games and  numerical results from arbitrary length games have provided strong indications that the optimal penalties (from the adversary's viewpoint) quickly become binary and remain so for the rest of the game. However, given the assumption that the sum of penalties
equals one, an additional binary penalty assumption implies that only one penalty component is equal to one, while all other components remain equal to zero. 
Effectively, the adversary is able to (maximally) punish exactly one option in each round.

We have shown in \cite{anagnostou2014playing} that a binary penalty assumption results to (a) a greedy optimal penalty scheme for the adversary, which further implies that (b) the optimal scheme ends up in being a rotating scheme (for all rounds $t$, such that $t>t_0$ for some $t_0$). The optimal scheme for the adversary is to 
assign a penalty equal to one to the option with the highest bet. In a set of initial rounds 
this strategy results into a near equalization of the bet distribution (over the options), since in each round $t$ the adversary always inflicts a unitary penalty on the option with the maximum  $w^t_i$ ($i=1,\ldots,N$), effectively driving Hedge to push the option $i$ weight downwards. Due to the quantized nature of the weight adjustments, the bet components (weights) never become exactly equal, but they become almost equal to each other and to $1/N$ within a margin. 
Then in each round the currently 
maximum weight is pushed below all other weights. 
Moreover, in this second phase the greedy adversarial algorithm effectively becomes a rotating scheme. 

\section{Rotating schemes}
\label{sec:rotationmotivation}
This is a brief section on rotating schemes, which serves as a preparation for the analysis of games that are long enough so as to include a rotational phase in their optimal adversarial scheme.

\subsection{Periodic penalties will produce periodic weights and periodic losses}
While the Hedge algorithm aids  the player to adapt the bet mixture so as to avoid the penalties imposed by the adversary, the latter may try to rotate the penalties, so as to make  adaptation difficult. In this section we explore the dynamics of Hedge under rotating penalty schemes. Hedge responds to a penalty rotation with weight rotation, as shown by the following lemma, which is valid for an $N$ option game:
\begin{lemma} \label{lemma:1}
Assume a rotating penalty scheme
\begin{equation}
\ell_{i+1}^{t}= \ell_i^{t-1} \quad (i=1, \ldots, N-1), \quad \ell_1^t = \ell_N^{t-1}
\label{periodicassumption2}
\end{equation}
i.e. option $i+1$ at time $t$  inherits the penalty of the previous option  $i$ in the previous round $t-1$.
Then the response of Hedge generates a loss in each round that is also periodic, i.e. it repeats itself after $N$ rounds.~$\Box$
\end{lemma} 

Note that (\ref{periodicassumption2}) implies that penalties repeat themselves after $N$ rounds:
\begin{equation}
\ell_i^{N+t}= \ell_i^{N } \quad (i=1, \ldots, N) \quad (t=0, \ldots, T-N)
\label{periodicassumption1}
\end{equation}
and in general 
\begin{equation}
\ell_i^{\tau}= \ell_{\tau + i    \bmod N}^{0}
\label{periodicassumption3}
\end{equation}
The proof of the above lemma is straightforward: Since $w_i^{t+1} = w_i^t \beta^{\ell_i^t}$
\[
w_i^{t+N}=w_i^t \beta^{\sum_{k=0}^{N-1} \ell_i^{t+k}} = w_i^t \beta^{\sum_{k=0}^{N-1} \ell_{t+i+k \bmod N}^{0}} 
= w_i^t \beta^{\sum_{k=0}^{N-1} \ell_{k}^{0}} = w_i^t \beta^{\Lambda} 
\]
where $\sum_{k=0}^{N-1} \ell_{k}^{0} = \Lambda$ is the sum of all the penalties in the initial or, due to the rotation, in any other round, 
and due to the normalization
\[
p_i^{t+N}= p_i^t 
\] 
In fact,  a periodic scheme would prevent Hedge from focusing on a single option. In Example~\ref{example:Noptionsexample} (see Appendix~\ref{appendix1}) the net loss performance is calculated 
for binary penalties.


\subsection{Periodic weights for maximum  loss}
\label{sec:optimalperiodicweights}
As explained in section~\ref{sec:rotationmotivation}, the adversary may use proper penalties with an aim to drive Hedge towards weights that will maximize loss in the rotation phase. Currently we calculate the optimal weights without exploring how  the adversary might possibly force Hedge to adopt these weights. The latter issue is examined later in this paper. 
 
We simplify our calculations by exploring a two option game.  Let $L_p$ denote the total loss per cycle, i.e. the cumulative loss in two consecutive rounds. Assuming that periodic behavior will start at $t_0$, which we set equal to 0 without loss of generality, let $(w,1-w)$ the pair of weights at $t=t_0$.
Assuming (also without loss of generality) that $w \geq 1/2$ (which also implies that $w \geq 1-w$), the adversary will choose penalties
$(\ell_1^{t_0}, \ell_2^{t_0}) = (1,0)$,
therefore the new pair of weights will be
\[
\left( \frac{w \beta}{w \beta + 1-w},  \frac{1-w}{w \beta + 1-w} \right)
\]
In the next round the adversary will choose penalties
$(\ell_1^{t_0+1}, \ell_2^{t_0+1}) = (0,1)$,
and the new pair of weights will be $(w,1-w)$ again.
The cumulative gain of the adversary (loss of the player) in both rounds will be
\[
L_p(w) = w + \frac{1-w}{w \beta + 1-w}
\]
which is maximized w.r.t. $w$ for 
\[ 
w = \frac{1}{ 1+ \sqrt{\beta}}
\]
Therefore the adversary would like to start with a pair of weights equal to
\begin{equation}
\left( \frac{1}{ 1+ \sqrt{\beta}} , \frac{\sqrt{\beta}}{ 1+ \sqrt{\beta}} \right)
\label{idealpairofweights}
\end{equation}
and by imposing a pair of penalties $(1,0)$ the pair of weights 
\[
\left(  \frac{\sqrt{\beta}}{ 1+ \sqrt{\beta}}, \frac{1}{ 1+ \sqrt{\beta}}  \right)
\]
will appear in the next round, and again the adversary will return to the original pair by using penalties $(0,1)$. The cumulative loss in these two rounds will be 
\[
L_p = \frac{2}{ 1+ \sqrt{\beta}}
\]
which for the adversary amounts to a constant gain $1/(1+ \sqrt{\beta})$ per round.

The pair $(1/(1+ \sqrt{\beta}),\sqrt{\beta}/(1+ \sqrt{\beta}))$ is for the adversary an attractive pair of weights given a sufficient time horizon $T$, so that this rotational steady state can be reached. However these values can in general be achieved by the adversary only by using fractional (i.e. non binary) penalties in at least one round. 

A straightforward generalization for $N$ options gives the optimal weights as
\[
w_i^*=\frac{(1-\beta^{\frac{1}{N}}) \beta^{\frac{i-1}{N}}}{1-\beta}, \quad i=1,\ldots,N
\]
and the loss per period of $N$ rounds is equal to
\[
\frac{N(1- \beta^{1/N})}{1-\beta}
\]

\section{Solutions with binary penalties}
It is possible to find a suboptimal solution by constraining the solution to binary penalties and using a greedy algorithm. In this section we summarize certain results of previous work  \cite{anagnostou2014playing}.

First, we reiterate the main idea in the analysis of binary penalties: If the current weights are $(w,1-w)$, a pair of penalties $(1,0)$ would transform the weights to 
\[
\left( \frac{w \beta}{w \beta + 1-w},  \frac{1-w}{w \beta + 1-w} \right),
\]
while penalties $(0,1)$ would produce weights equal to
\[
\left( \frac{w }{w  + (1-w)\beta},  \frac{(1-w)\beta}{w + (1-w) \beta} \right)
= \left( \frac{w \beta^{-1}}{w \beta^{-1} + 1-w},  \frac{1-w}{w \beta^{-1} + 1-w} \right)
\]
Let us define 
\begin{equation}
f(w,x) \equiv \frac{w \beta^{x}}{w \beta^{x} + 1-w}, 
\label{defn:fx}
\end{equation}
but we drop $w$ whenever it is obvious and write $f(x)$.
By using this notation,
in one round the first option weight moves from $f(0)=w$ either to $f(\beta)$, or to $f(-\beta)$. In $n$ rounds, if the number of first option penalties that are equal to~1 is $k$, the weights before normalization are $(w \beta^k, (1-w) \beta^{n-k})$, therefore the first option weight is
\[
\frac{w \beta^k}{w \beta^k +  (1-w) \beta^{n-k}} = \frac{w \beta^{2n-k}}{w \beta^{2n-k} +  1-w} 
\]
Effectively, each new round brings a move on $f(x)$ such that $x$ increases or decreases by one (penalty) unit. In each move the adversary has the option to increase the player's loss (i.e. the adversary's gain) by $f(x)$ by moving to $x+1$, or increase it by $1-f(x)$ and move to $x-1$. The rest of the analysis in \cite{anagnostou2014playing} explains why this walk on the curve $f(x)$ can be optimal for the adversary by making greedy choices (i.e. move to $x+1$ if $f(x) \geq 1/2 \geq 1-f(x)$, otherwise move to $x-1$). 
The main idea in this proof is that each forward move (towards $x+1$) is for the adversary a high gain move if $f(x) \geq 1/2 \geq 1-f(x)$, but it decreases the weight, thus it also decreases the possible gain in the next moves. However, we have proven that any attempt of  the adversary to ``invest in the future'', by currently accepting a lower gain (equal to $1-f(x)$ instead of $f(x)$) and by moving backwards, will never be able to produce adequate future gains that will justify the current sacrifice.  This result is summarized in the following lemma:
\begin{lemma}
The optimal solution with binary penalties is greedy.
\end{lemma}

\section{How to calculate the optimal penalties}
\label{sec:optimalpenalties}
\subsection{Introductory remarks and an example}
Firstly we give some numerical results using expressions (\ref{eq:TgamesTotalLossRecursionVect}), 
(\ref{eq:optimalLastPenaltyVect}), and 
(\ref{eq:optimalOtherPenaltiesVect}). 

\begin{example}
\label{example:balancedgame}
We set $\beta = 0.8$ and $T=10$ and produce the total cumulative loss numerically by using formulas \ref{eq:TgamesTotalLossRecursionVect}, \ref{eq:optimalLastPenaltyVect}, and \ref{eq:optimalOtherPenaltiesVect}. We have quantized the possible values of the (first option) initial weight $w$ to values equal to $i/10000$, where $i=0,1,2\ldots, 10000$; therefore the total cumulative losses, the penalties etc as functions of $w$ are vectors of 10001 components. 
The resulting total cumulative loss vs $w$ is shown on Fig.~\ref{fig:numericalMultipleRoundParameterized4fig3}, while Fig.~\ref{fig:withverticallines} shows the (first option) penalties of different rounds that have led to this result. Let us summarize a few initial observations:
\begin{enumerate}
\item As expected, there is an even symmetry for the total cumulative loss, i.e. 
$L^{T-1}_{\max}(w)=L^{T-1}_{\max}(1-w)$ and an odd symmetry for the associated penalties, i.e. $\ell_1^t(w)=1-\ell_1^t(1-w)$.
\item Non binary penalty values occur mostly in the first round. There is also an occurrence in the second round, but not in both rounds in the same game. 
The non binary values are limited to   distinct small areas of $w$ in the first round. Both the lengths of these areas and the aberrations from the binary values are quite limited.
Some of these areas have been marked with vertical lines at both ends.
\item For  $w$  very close to 0 or 1 all penalties are equal to~0 or~1 respectively, while for $w$ very close to $1/2$ there is a pattern of alternating binary penalties.  Clearly, there is a first phase of high gains per round, in which the adversary tries to take maximum advantage of the largest weight by always penalizing the same option, and a second phase, in which the weights oscillate between two values that are close to $1/2$. Therefore, in this second phase the adversary cannot extract an average  per round gain significantly higher than $1/2$. The length of the first phase depends on the initial weight $w$ (and on $\beta$). The closer the value of $w$ is to $1/2$, the sooner the game enters the phase of rotation.
\end{enumerate}

\begin{figure}
	\centering
  \includegraphics[width=.6\textwidth]{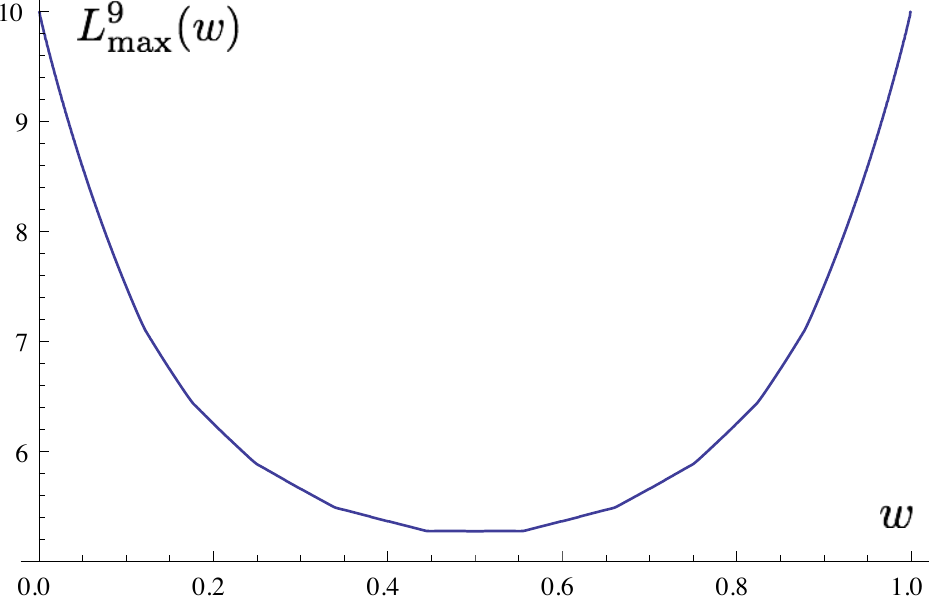}
  \caption{Total cumulative loss after $T=10$ rounds vs initial (first option) weight $w$  (for $\beta = 0.8$).}
  \label{fig:numericalMultipleRoundParameterized4fig3}
\end{figure}

\begin{figure}
	\centering
  \includegraphics[width=.6\textwidth]{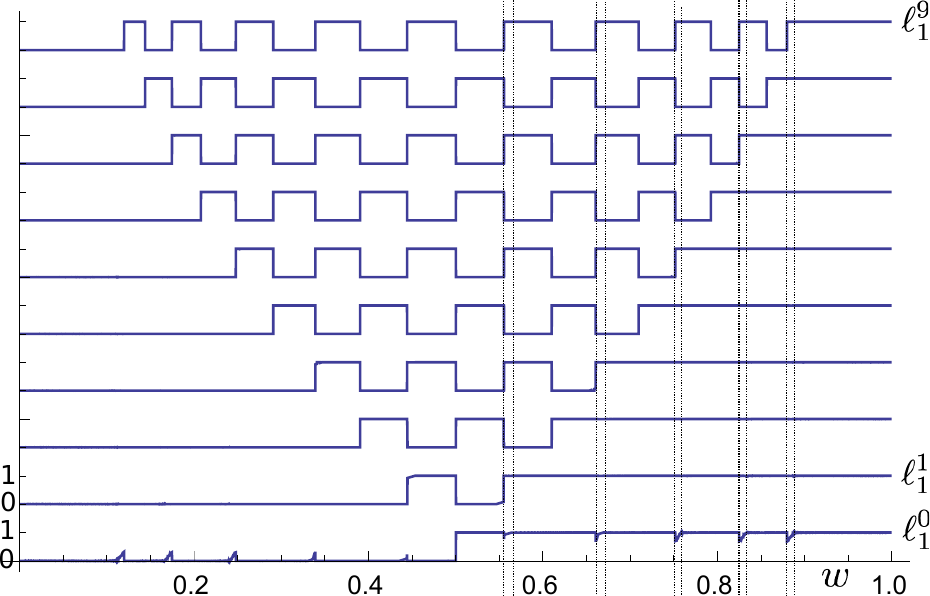}
  \caption{Round $i$ first option penalty $\ell_1^i$ vs initial (first option) weight $w$ (for $\beta = 0.8$).}
  \label{fig:withverticallines}
\end{figure}

We name the first phase ``the transitional phase'', and the second phase ``the rotational phase'', and we define the precise limit between the two phases later on.~$\Box$
\end{example}

Assuming a weight $w$ (for the first option) at some round, if the adversary chooses a penalty $x$ such that $0 \leq x \leq 1$ (for the first option, and $1-x$ for the second option), the player's loss in this round will be equal to $wx+(1-w)(1-x)$, and the new (first option) weight that will appear in the next round is equal to 
\[
\frac{w \beta^x}{w \beta^x + (1-w)\beta^{1-x}} = \frac{w \beta^{2x-1}}{w \beta^{2x-1} + 1-w} 
\]
In each round the adversary can choose any value of $x$ between~0 and~1, but (assuming that $w>1/2$) the new weight will be smaller or greater than $w$ depending on whether $x>1/2$ or $x<1/2$ respectively, and will remain the same if $x=1/2$.

In general, if the sum of first option penalties is equal to $x$ in the next $n$ rounds, the pre-normalized weights will be $(w \beta^x, (1-w) \beta^{n-x})$, and the final first option weight will be
\[
\frac{w \beta^x}{w \beta^x + (1-w) \beta^{n-x}} = \frac{w \beta^{2x-n}}{w \beta^{2x-n} + 1-w} = f \left( 2x-n \right),
\]
where $f(x)=f(w,x)$ is defined by (\ref{defn:fx}). 
\subsection{The transitional phase}
We use the above observation to prove some useful minor theorems.
The first one describes a situation, in which the initial weights $(w_1^0, w_2^0)=(w,1-w)$ are unequal, and their relative size cannot be reversed in the next $n_1$ rounds, i.e. $w \beta^{n_1} > 1-w$ ($n_1 \leq  \lfloor  \ln \frac{1-w}{w} / \ln \beta \rfloor $). Under this assumption we show that in a game with $T \leq n$ rounds the optimal strategy of the adversary is the greedy strategy, i.e. to set all first option penalties to~1, instead of using any fractional penalties. First, we give the following lemma:
\begin{lemma} 
\label{lemma:phaseonegameonly}
If $0<\beta<1$, $w > 1/2$, $n$ is such that $w \beta^{n+1} > 1-w$, and $0 < \epsilon \leq 1$, then for $f(x)$ (as defined by (\ref{defn:fx})) the following inequality is valid:
\begin{equation}
( 1- \epsilon ) f(0) + \epsilon \, [1-f(0)] + \sum_{k=1}^{n} f(k-2 \, \epsilon) < \sum_{k=0}^{n} f(k) 
\label{ineq:dilemma2}
\end{equation}
\end{lemma}
Effectively, in a game of $T$ rounds 
starting with $w_1^0=w$ such that $w>1/(1+\beta^T)$
(i.e. the number of rounds is not sufficient to reverse the balance of weights between the two options  after the end of the game, even if the first option is always maximally penalized)%
, the optimal  policy of the adversary is the greedy one, i.e.   to use unitary penalties  in all rounds ($\ell_1^i=1$ for $i=0,1,\ldots, T-1$).
Although the first part of this lemma strictly states that a non binary penalty should not be chosen  in the initial round, it can also be used to exclude a non-binary step in any intermediate round if we assume that the game starts exactly at this round. In addition, it will be shown in a later lemma that the earlier a ``sacrifice'' (i.e. a deviation from the greedy policy, which gives a maximum short term gain to the adversary) is, the more effective it is for the adversary. Therefore any sacrifice should be undertaken in the very first round.
The above lemma can be proved by induction. However, we have omitted the proof.

\begin{figure}[htb]
	\centering
  \includegraphics[width=.7\textwidth]{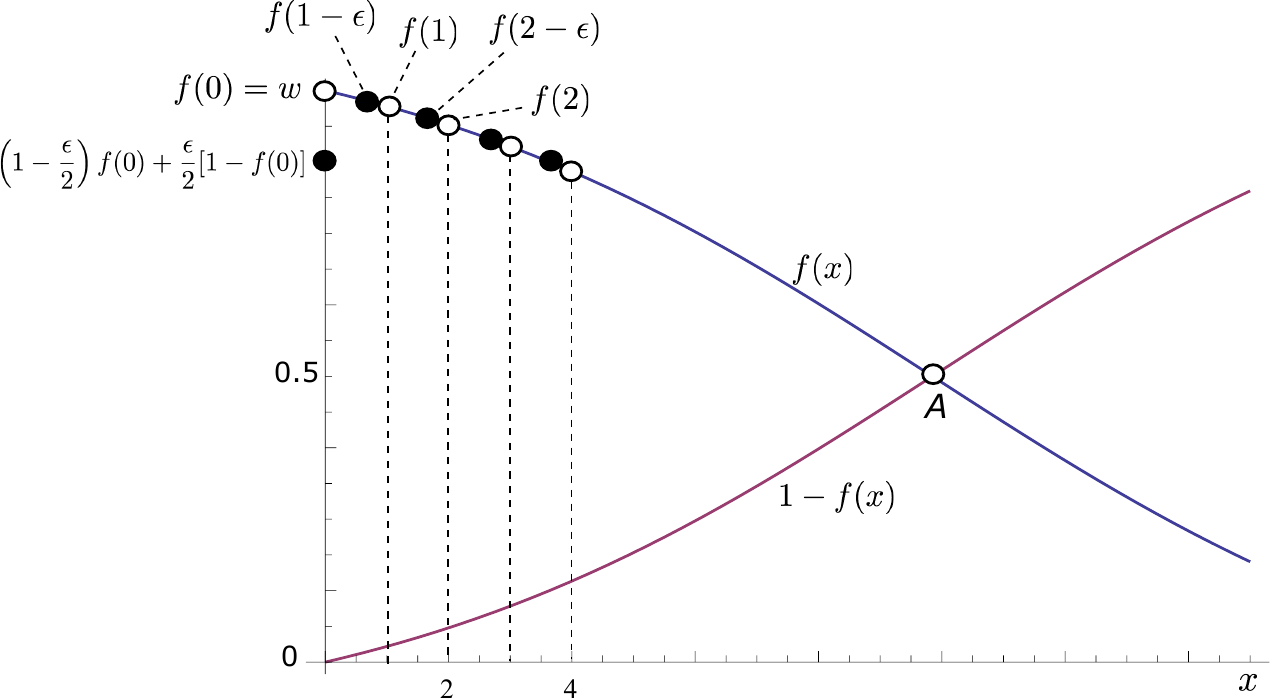}
  \caption{Interpretion figure for Lemma~\ref{lemma:phaseonegameonly}.}
  \label{fig:fofx}
\end{figure}

We shall give an interpretation of the above lemma and a few additional details. Consider Fig.~\ref{fig:fofx}.
The initial weights are $(w_1^0, w_2^0)=(w,1-w) = (f(0), 1- f(0))$. If the adversary implements the greedy strategy continuously for the next $n$ rounds,
the first option weights are
$f(x_0), f(x_1), \ldots f(x_n)$, where $x_0=0$, and $x_{k+1}=x_k+ 2 \ell^k_{1} = x_k + 2 \times 1 - 1 = x_k - 1$. Therefore, under the greedy strategy the first option weights are $f(0), f(1), \ldots, f(n)$ (i.e. the white dots) and the player's total loss (in $n+1$ rounds) is $f(0) + f(1) + \ldots + f(n)$. On the other hand, the adversary could choose to make a sacrifice in the first round, so as to imcrease loss in the remaining rounds as shown by the black dots. Effectively, Lemma~\ref{fig:fofx} states that the sacrifice in the first round cannot be counterbalanced by the improved weights in all subsequent rounds. Note, however, that Lemma~\ref{fig:fofx} holds if the initial weight $w$ is such that $w \beta^T > 1-w$, i.e. if the weight that would appear in a $T+1$'th round (that does not exist) would still be above $1/2$.

If $w>1/2$, but $w \beta^{T} < 1-w$, the area in which rotational schemes are possible is reachable by the adversary by continuously using penalties equal to~1 until the weight of option~1 falls below $1/2$. In the rest of the game the adversary could use an optimized rotation scheme. In order to reach this optimized rotation the adversary may need to make a fractional penalty sacrifice at some point. Assuming that such a non greedy step should be taken, the question of its timing emerges. Should the non greedy step be taken just before entering rotation or perhaps earlier? A lemma that will soon be introduced states this step should be executed as early as possible.

Before dealing with the problem of optimally entering the rotation phase, we shall briefly examine a marginal case, in which the weight of the final round gets so close to $1/2$, that it would fall below $1/2$ if  an additional round existed. Effectively, this implies for the initial weight that $w \geq 1/(1+\beta^{T-1})$, but $w<1/(1+\beta^{T})$, therefore Lemma~\ref{lemma:phaseonegameonly} would not be valid. The optimal policy of the adversary is to make a non binary adjustment in the first round, which will improve the gains of the greedy moves in all subsequent rounds. Therefore, in this marginal situation it can be proved that the optimal set of penalties is $(\ell_1^0, \ell_1^1, \ldots, \ell_1^{T-1}) = (1-\epsilon, 1, \ldots, 1)$. The correct value of $\epsilon$ can be found by using direct optimization of the resulting single variable objective function for the total cumulative loss, and it approaches zero as $w$ moves from $1/(1+\beta^{T})$ towards $1/(1+\beta^{T-1})$. 
In fact, extensive numerical tests provide strong indication that there is a marginal value $w_m \in (1/(1+\beta^{T-1}), 1/(1+\beta^{T}))$ such that the optimal $\epsilon$ is zero if $w \geq w_m$, but it is non zero as $w$ becomes lower than $w_m$ and it grows linearly as $w$ approaches the lower end of this area, i.e. $1/(1+\beta^{T-1})$.
Numerical tests also show that for $w=1/(1+\beta^{T-1})$ the optimal $\epsilon$ is around $1/4$ regardless of the value of $\beta$. Effectively this means that given an initial weight $w$ in the interval $1/(1+\beta^{T})$ towards $1/(1+\beta^{T-1})$, the adversary will need to maximize the total cumulative loss w.r.t. $\epsilon$, and the expected outcome is between zero and $1/4$.
However, the analysis given in the above few lines is probably too detailed and all that is necessary is the following lemma, which is given without proof:
\begin{lemma}
\label{lemma:marginal}
If $  1/(1+\beta^{T-1}) \leq w<1/(1+\beta^{T})$, there is an $\epsilon$, such that $0 < \epsilon \leq 1$, for which the optimal policy of the adversary is achieved by setting $\ell_1^0 = 1 -\epsilon$, and $\ell_1^i = 1$ in the remaining rounds $i=1,\ldots,T-1$.~$\Box$
\end{lemma}

Effectively, if the assumptions of Lemma~\ref{lemma:marginal} are true, the inequality (\ref{ineq:dilemma2}) may be reversed for some $\epsilon >0$. 
Note, however, that the weight $w^{T-1}_1$ that enters the last round is still above $1/2$, therefore Lemma~\ref{lemma:phaseonegameonly} and Lemma~\ref{lemma:marginal} are separated by a very thin line.
Both lemmas assume that the (first option) weight will never fall below $1/2$ at any round, but Lemma~\ref{lemma:marginal} states that special care is needed in the first round if the final round weight is likely to come very close to $1/2$.

The last lemma of this section states (as already promised) that if a non binary penalty must be used, it should appear as early as possible. The lemma compares two scenarios: In the first scenario the fractional step precedes a sequence of $n$ successive unit penalty steps, while in the second scenario the fractional step follows the unit steps. The loss in the first scenario (with an analysis similar to the analysis of Lemma~\ref{lemma:phaseonegameonly}) is $f(0)(1-\epsilon)+(1-f(0))\epsilon$ in the first round, and $f(k - \epsilon)$, ($k=1,2,\ldots,n$) in the following $n$ rounds. This gives a total loss equal to
\[
f(0)(1-\epsilon)+(1-f(0))\epsilon + \sum_{k=1}^n f(k -\epsilon)
\]
In the second scenario the fractional penalty is used in the last round, thereby giving a total loss equal to
\[
\sum_{k=0}^{n-1} f(k)+ f(n)(1-\epsilon)+(1-f(n))\epsilon .
\]
The proof of the lemma can  be produced by induction, starting from two rounds ($n=1$).
\begin{lemma} 
\label{lemma:targetweightaboveinters}
If $0<\beta<1$, $w > 1/2$, $n$ is such that $w \beta^n > 1-w$, and $0 < \epsilon \leq 1$, then for $f(x)$ (as defined by (\ref{defn:fx})) the following inequality is valid:
\[
\sum_{k=0}^{n-1} f(k)+ f(n)(1-\epsilon)+(1-f(n))\epsilon < f(0)(1-\epsilon)+(1-f(0))\epsilon + \sum_{k=1}^n f(k -\epsilon)  \quad \Box
\]
\end{lemma}
Note that both scenarios of Lemma~\ref{lemma:targetweightaboveinters} produce the same final weight $w^{n+1}_1=f(n+2-2 \epsilon)$; this is not true for the previous lemmas in this section. The final weight $f(n+2-2 \epsilon)$ is greater than $1/2$ due to the assumption $w \beta^n > 1-w$.
The final weight could serve as a target weight; for example it could be set equal to the ideal weight $1/(1+\sqrt{\beta})$ or some other suitable value. Lemma~\ref{lemma:targetweightaboveinters} states that the non binary penalty step should be taken as early as possible.

We are now about to examine games that are long enough to produce (first option) weights lower than $1/2$.
Before stating any additional lemmas we define the ``intersection area'': This is an area of weights around $1/2$, and  includes any weight from which $1/2$ is reachable in a single round. Effectively this implies a penalty equal to 1, if the starting weight is greater than $1/2$, or a penalty equal to 0, if the starting weight is lower than $1/2$. Therefore the upper limit of this area is a weight $u_1$, such that 
$u_1 \beta / (u_1 \beta + 1-u_1) = 1/2$, which gives $u_1=1/(1+\beta)$, and the lower limit is $u_2$, such that $u_2/(u_2+(1-u_2)\beta)=1/2$, which gives $u_2=\beta / (1+ \beta)$. Thus the ``intersection area'' is the interval $[\beta / (1+ \beta), 1 / (1+ \beta)]$.This area obviously includes the optimal rotational weights $\sqrt{\beta}/(1+\sqrt{\beta})$ and $1/(1+\sqrt{\beta})$, which are reachable from one another with binary penalties.

\subsection{Entering the rotational phase} 
If $w \beta^{T-1} > 1-w$ the adversary cannot reach a weight below $1/2$ ever in the game. According to the previous lemmas, if the initial weight is somewhat greater, i.e. it is such that $w \beta^{T} > 1-w$ (which implies $w \beta^{T-1} > 1-w$), then the optimal policy of the adversary is to use a sequence of penalties equal to $(1,  1, \ldots, 1)$. 
If, however, $w \beta^{T-1} > 1-w$ holds marginally so that  $w \beta^{T} < 1-w$, the adversary might benefit from using a non binary first round penalty. 
Finally, if $w \beta^{T-1} < 1-w$, a penalty sequence equal to $(1,  1, \ldots, 1)$ will eventually bring the weight below $1/2$. In the rest of this paper we shall see that the adversary's optimal plan is roughly to bring the current weight ($p_1^t$) as fast as possible from the initial weight $w=p_1^0$ to a value close to $1/2$, and then produce weights that rotate around $1/2$ (by using a sequence of alternating 0 and 1 penalties). However, as devil is in the details, we shall also see that there is the issue of optimally approaching the most suitable rotating weights, while both the optimal approach and the most suitable weight values depend on the total length $T$ of the game. 
All these issues will be explored  by using a number of lemmas.

The optimal weights are not necessarily those given by (\ref{idealpairofweights}). 
Assume  that the latest weight achieved by a so far greedy adversary is $u$ and that $u$ is about to enter the intersection area, in the sense that in two rounds the weight could plunge below 1/2, i.e. it is such that $u \beta^2 < 1-u$, although $u \beta > 1-u$ (i.e. one round is not enough to reach $1/2$). Suppose also that somehow the adversary has computed $w'$ as the ideal weight for the future, and intends to approach $w'$ as soon as possible.
 For example, under certain conditions the ideal weight $w'$ could be equal to $1/(1+\sqrt{\beta})$.
 If by any chance $u \beta /(u \beta + 1-u) = w'$, then the adversary can (still be considered ``lucky'' and) set the next penalty equal to~1 and reach $w'$ in one round. If $u \beta /(u \beta + 1-u) = w'-\epsilon$ for some small $\epsilon$, the adversary can still consider himself ``lucky'': 
 He  can use a penalty $1-\delta$ for some small suitable $\delta$, and reach the desired weight by sacrificing a small part of the gains. 
 If, however,  $u \beta /(u \beta + 1-u) = w'+\epsilon$, the weight $w'$ is unreachable in just one round. The adversary could possibly use penalties $(1/2+ \delta , 1)$ or $(1, 1/2+ \delta)$, in order to reach $w'$, but a  penalty equal to $1/2+\delta$ is implies a significant sacrifice (as compared to a penalty equal to~1 applied on the greatest weight). 
 A better solution is to let the first option weight fall below $1/2$ by applying a penalty vector $(1,1)$ in the next two rounds, and then approach the target weight from below by penalizing the now stronger second option weight with a penalty $\ell_2=1-\delta$. The next two lemmas state the appropriate policy when the adversary is about to approach to a target weight $w'$.

\begin{figure}[htb]
	\centering
  \includegraphics[width=.7\textwidth]{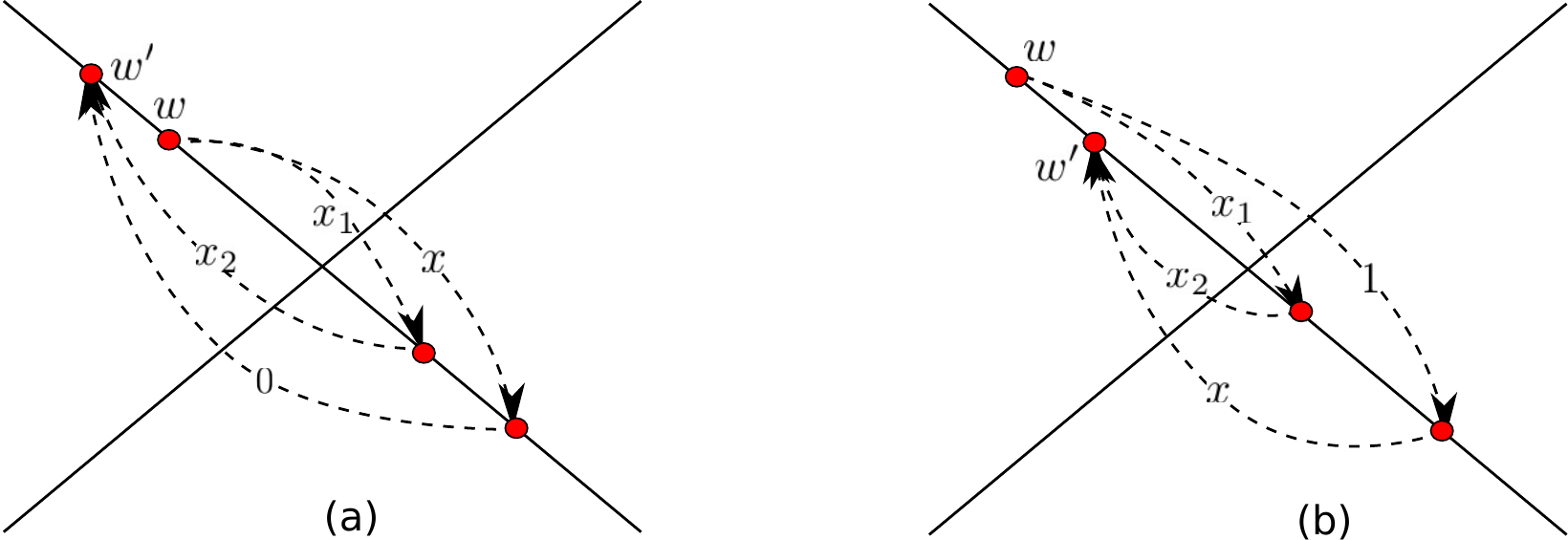}
  \caption{Interpretion figure for Lemmas~\ref{lem:rotation1} and \ref{lem:rotation2}.}
  \label{fig:rotationlemma1}
\end{figure}

\begin{lemma}
\label{lem:rotation1}
Assuming that a game starts with $w$, and $w'$ is a target weight such that $1/2< w \leq w' < 1/(1+\beta)$, and the adversary's intention is to achieve $w'$ within two rounds, and $x, x_1, x_2$ are such that $f(x-2)=w'$, $x=x_1+x_2 \Rightarrow f(x_1+x_2-2)=w'$, $0 < x_1< x \leq 1$, $0< x_2 \leq 1$, then the adversary should prefer the penalty vector $(\ell^0_1, \ell^1_1) = (x,0)$ over the (less greedy) penalty vector $(x_1,x_2)$. In other words
\begin{eqnarray*}
&& w x + (1-w)(1-x) + \frac{(1-w) \beta^{1-x}}{w \beta^x + (1-w) \beta^{1-x}} > \\
&&
w x_1 + (1-w)(1-x_1) \\
&+& x_2   \frac{w \beta^{x_1} }{w \beta^{x_1} + (1-w)\beta^{1-x_1}} + (1-x_2) \frac{(1-w)\beta^{1-x_1} }{w \beta^{x_1} + (1-w)\beta^{1-x_1}}. \quad \Box
\end{eqnarray*}
\end{lemma}
This situation is shown in Fig.~\ref{fig:rotationlemma1}a. Both weights $w$ and $w'$ are above $1/2$, and both are in the intersection area. The lemma states that the adversary should go for the maximum possible penalties $(\ell^0_1, \ell^1_1)$ in each round; this implies $\ell_2^1=1-\ell_1^1=1$ in the second round, and the maximum possible penalty in the first round, such that $w'$ will be achieved after two rounds. The proof is trivial, since in both rounds the preferred policy $(x,0)$ brings a higher gain than the policy $(x_1,x_2)$. Next comes the lemma complementary to the previous one for $w> w'$, as shown in Fig.~\ref{fig:rotationlemma1}b.

\begin{lemma}
\label{lem:rotation2}
Assuming that a game starts with $w$, and $w'$ is a target weight such that $1/2< w' \leq w < 1/(1+\beta)$, and the adversary's intention is to achieve $w'$ within two rounds, and $x, x_1, x_2$ are such that $f(x-1)=w'$, $1+x=x_1+x_2 \Rightarrow f(x_1+x_2-2)=w'$, $0< x_1 < 1$, $0 < x_2< x \leq 1$, then the adversary should prefer the penalty vector $(\ell^0_1, \ell^1_1) = (1,x)$ to the (less greedy) penalty vector $(x_1,x_2)$. In other words
\begin{eqnarray*}
&& w  + x \frac{w \beta}{w \beta +1-w}  + (1-x) \frac{1-w}{w \beta +1-w} > \\
&&
w x_1 + (1-w)(1-x_1) \\
&+& x_2   \frac{w \beta^{x_1} }{w \beta^{x_1} + (1-w)\beta^{1-x_1}} + (1-x_2) \frac{(1-w)\beta^{1-x_1} }{w \beta^{x_1} + (1-w)\beta^{1-x_1}}. \quad \Box
\end{eqnarray*}
\end{lemma}

However, the value of the aforementioned sacrifice (or investment) has also to be weighed against 
the expected gains from the target weights, whose values depend on the remaining rounds. The more the remaining rounds are, the better the justification of approaching the ideal rotational weights is. 

\begin{example} 
\label{example:4variantspart1}
Assume $\beta = 0.8$,  $w_1^0=w=0.5108$ and a very large number of rounds $T$. In this case the adversary would pursue the pair of weights $((1/(1+\sqrt{\beta}),\sqrt{\beta}/(1+\sqrt{\beta}) = (0.5279, 0.4721)$ for the rotational phase. The penalty $\ell_1^0=0.8469$ brings the weight to  $w_1^1=\sqrt{\beta}/(1+\sqrt{\beta}) = 0.4721$ and the next penalty $\ell_1^1=0$ to $w_1^2=1/(1+\sqrt{\beta}) = 0.5279$, in accordance with Lemma~\ref{lem:rotation1}.

In another variant of the same example, assume that  $w_1^0=w=0.6200$. The adversary could reach the target weight in two rounds, while the penalty vector $(1,1)$ would bring the weight $w_1^2$ to the same as above value $0.5108$, which means that a penalty vector $(1,1,0.8469)$ would bring the weight $w_1^3$ to the target value $0.4721$ in three rounds. However, since the adversary now has a margin of 3 rounds to properly reach the target weight, it is possible to perform the adjustment in a previous round by opting for penalties $(1,0.8469,1)$ or $(0.8469,1,1)$. In fact the best option is to implement the adjustment in the initial round, i.e. to choose $(0.8469,1,1)$. While the cumulative loss in the first three rounds is $1.6937$ for $(1,1,0.8469)$, it is $1.6940$ for $(0.8469,1,1)$. Of course, the rest of the loss is the same in the subsequent rounds, i.e. $(1/(1+\sqrt{\beta})=0.5279)$ per round for the remaining $T-3$ rounds of the game.

In yet another variant, assume now that $w_1^0=w=0.6200$ as in the second variant, but the game length is now restricted to $T=10$ rounds. While for the whole game the initial round adjustment (and then greedy) penalty vector $(0.8469,1,1,0,1,0,1,0,1,0)$ produces a total cumulative loss equal to $5.38908$ and is still superior to the third round adjustment $(1,1,0.8469,0,1,0,1,0,1,0)$ vector, which produces a loss equal to $5.38876$, the strictly greedy vector $(1,1,1,0,1,0,1,0,1,0)$ is even better as it produces a loss equal to $5.40886$.  
The per round average loss of the greedy solution in the rotational phase is inferior to the loss of the adjusted solution ($0.52783$ vs $0.52786$), but the adjusted solution  cannot counterbalance the sacrifice in the first round (first round loss equal to $0.58326$ in the adjusted solution instead of $0.62000$ in the greedy solution). 
This explains why the optimal penalties in Fig.~\ref{fig:numericalMultipleRoundParameterized4fig2} remain greedy around $w=0.62$.
Even direct optimization of the loss produced by the penalty vector $(1-\delta,1,1,0,1,0,1,0,1,0)$ w.r.t. $\delta$, produces the solution $\delta=0$ and confirms the previous considerations

In a final variant, assume that $w=0.65$ ($\beta=0.8$), and $T$ is large. After two greedy rounds with penalties $(1,1)$
the current weight $w_1^2$ is at $0.5431$, still above the target $0.5279$, but close.
The mirror target $1-0.5279$ is out of reach in one round.
Although the adversary could use a penalty equal to $0.1371$ and reach $0.5279$, that would be a bad decision, since  the situation is more like Fig.~\ref{fig:rotationlemma1}. 
The target can be approached from below in two rounds, by using penalties $(\ell^2_1, \ell^3_1) = (1, 0.1371)$. Effectively the correct policy is $(1,1,1,0.1371,1,0,1,0,\ldots)$. However, for a short game with $T=10$ rounds the greedy policy $(1,1,1,0,1,0,1,0,1,0)$ is preferable as in the previous variant.~$\Box$
\end{example}

In the above lemmas and examples the issue of the proper way of entering the rotational phase was explored. On occasion this involves a round, in which an non binary penalty adjustment must be made. If the current weight is $u$ and the target weight is $u'$, it is easy to see that the penalty $x$ that achieves the transition from $u$ to $u'$ in one round is 
\begin{equation}
x(u,u')= \frac{1}{2} + \frac{\ln \left( \frac{u'(1-u)}{u(1-u')} \right)}{2 \ln \beta}
\label{properpenalty}
\end{equation}
assuming that $u'$ is reachable from $u$ in a single round, i.e. 
$u \beta / (u \beta +1-w) \leq u'$.

If the adversary has chosen a greedy policy in the first phase, assuming (sufficiently unequal) initial weights $(w, 1-w)$, the possible (first option) weight values from which to make a non binary non greedy adjustment are $w_1^n= w \beta^n /(w \beta^n + 1-w)$.
If, in addition, the target weight is $w_1^*=1/(1+\sqrt{\beta})$, then 
\[
x(w_1^n,w_1^*)=\frac{1}{4} - \frac{n}{2} + \frac{\ln \left( \frac{1-w}{w} \right)}{2 \ln \beta}
\]
Since $0 \leq x(w_1^n,w_1^*) \leq 1$, the integer $n$ must satisfy
\[
- \frac{3}{2} + \frac{\ln \left( \frac{1-w}{w} \right)}{\ln \beta} \leq n \leq
\frac{1}{2} + \frac{\ln \left( \frac{1-w}{w} \right)}{\ln \beta}
\]
If the target is $w_2^*= \sqrt{\beta} /(1+\sqrt{\beta})$
\[
x(w_1^n,w_2^*)=\frac{3}{4} - \frac{n}{2} + \frac{\ln \left( \frac{1-w}{w} \right)}{2 \ln \beta} = x(w_1^n,w_1^*) + \frac{1}{2}
\]
and $n$ must satisfy
\[
- \frac{1}{2} + \frac{\ln \left( \frac{1-w}{w} \right)}{\ln \beta} \leq n \leq
\frac{3}{2} + \frac{\ln \left( \frac{1-w}{w} \right)}{\ln \beta}
\]
If the adversary is in a greedy first phase, there are only two appropriate successive values of $n$ for any target weight $u'$ from the available weights $w_1^n= w \beta^n /(w \beta^n + 1-w)$. The first one is the maximum integer $m$ such that $w_1^{m}$ is still larger than $u'$ but $w_1^{m+1} \leq u'$, and the second one is $m+1$. Suppose also that $u'$ is such that $u'>1/2$, 
but $u''=u' \beta /( u' \beta +1-u) < 1/2$, i.e. $u'$ and $u''$ are likely to be used as weights in the rotational phase. Then the adversary
would possibly consider the penalty scenario
\[
(\ell_1^0, \ldots , \ell_1^{T-1}) = (\underbrace{1,1,\ldots, 1}_{m},x,1,0,1,0, \ldots),
\]
which approaches $u'$ from above, or the scenario
\[
(\ell_1^0, \ldots , \ell_1^{T-1}) = (\underbrace{1,1,\ldots, 1}_{m},1,x,0,1,0, \ldots),
\]
which approaches $u'$ from below, where $x$ is an appropriate non binary penalty.

Now we turn to the proble of choosing  the best binary correction.
Suppose that the initial weight $w$ is such that the first option weight could fall below
$1/2$ in exactly $r$ rounds, i.e. $w \beta^{r-1} > 1-w$, but  $w \beta^{r} \leq 1-w$.
The greedy penalty sequence is 
\[
(\underbrace{1,1,\ldots, 1}_{r},\underbrace{0,1,0, \ldots}_{T-r})
\]
If $T-r$ is even, the total cumulative loss under this sequence is
\[
\sum_{i=0}^{r-1} f(i) + (1-f(r)) \frac{T-r}{2} + f(r-1)  \frac{T-r}{2} ,
\]
and if $T-r$ is odd, the loss is
\[
\sum_{i=0}^{r-1} f(i) + (1-f(r)) \frac{T-r+1}{2} + f(r-1)  \frac{T-r-1}{2}.
\]

Taking into account the previous discussion, the adversary might try the sequence 
\[
(\underbrace{1,1,\ldots, 1,x, 1}_{r},\underbrace{0,1,0, \ldots}_{T-r}),
\]
assuming that $x$ is close enough to~1, so that $w \beta^{r-2+x} > 1-w$, but $w \beta^{r-1+x} \leq 1-w$ remains valid.
Then for even $T-r$ this sequence would create a total cumulative loss equal to
\begin{eqnarray*}
&&\sum_{i=0}^{r-3} f(i)+ x f(r-2)+ (1-x) [1-f(r-2)]+f(r+2x-3)\\
&+& [1-f(r+2x-2)] \frac{T-r}{2}+
f(r+2x-3) \frac{T-r}{2}
\end{eqnarray*}
However, in this case the adversary would rather pull the adjustment to the very first round by using the penalty sequence
\[
(\underbrace{x,1,1, \ldots,1}_{r},\underbrace{0,1,0, \ldots}_{T-r}),
\]
which yields the following total cumulative loss:
\begin{eqnarray}
&& x f(0) + (1-x) [1-f(0)]+ 
\sum_{i=1}^{r-1} f(2(x+i-1)-i) \nonumber \\
&+& [1-f(r+2x-2)] \frac{T-r}{2}+
f(r+2x-3) \frac{T-r}{2}
\label{eq:totallossasfofx}
\end{eqnarray}
While the most advantageous approach for the adversary depends on the relative position of $w^n_1$ and the target weight, as detailed in previous parts of this section, the maximum adjustment (value of $x$) is determined by the optimal rotational weights $w_1^*, w_2^*$. When the weight just before the adjustment is above $1/2$, $x$ is between 1 and the value that leads to  $w_1^*$ or $w_2^*$ (depending on the proximity to $1/2$). When the weight is below $1/2$, $x$ is between 0 and the value that leads to $w_1^*$. The longer the remaining game duration, the more likely is $x$ to try to approach the optimal rotational values, but if the remaining game is short, a binary value will be preferred. We can continue the previous example in order to show how this approach works.

\begin{example}
\label{example:4variantspart2}
In the second and third variant of Example~\ref{example:4variantspart1} for $w=0.62$, $\beta = 0.8$, the adversary has chosen the penalty sequence $(0.8469, 1, 1, 0, 1, 0,1, \ldots )$, which forces the weights to converge to the optimal rotational weights $w_1^*, w_2^*$ after the third round. However, it was
pointed out that a ten round game is not long enough to justify the sacrifice necessary for the adjustment, since the sequence $(1, 1, 1, 0, 1, 0,1, 0,1,0 )$ produces a definitely superior result. This fact indicates that perhaps there is an  optimal sequence $(x, 1, 1, 0, 1, 0,1, 0,1,0 )$, where $0.8469 \leq x \leq 1$.

The first round loss without adjustment is equal to $w \times 1 + (1-w) \times 0 = w = 0.62$. The first round loss with the aforementioned optimal rotational adjustment $x^* = 0.8469$ is $w x^* + (1-w) (1-x^*) = 0.5833$. Therefore the adversary's sacrifice when $x^*$ is used is equal to $0.0367$. On the other hand the optimal rotational loss per round is $w_1^*$, while the 
loss per round (in the second phase) in the greedy scenario is $[f(2)+1-f(3)]/2$. The difference $w_1^* - [f(2)+1-f(3)]/2 = 0.527864 - 0.527832 = 0.000032$ is so small that more than one thousand rounds are needed to balance the sacrifice.
Direct numerical optimization of the total cumulative loss w.r.t. $x$ (given by expression \ref{eq:totallossasfofx}) for a 643 round long game gives a value of the optimal $x$ equal to $0.8583$, which shows convergence towards $x^* = 0.8469$ as $T$ increases.~$\Box$
\end{example}

\subsection{A last micro-adjustment before entering rotation} 
\begin{example}
\label{example:microadjust}
Let us re-examine the optimal penalties of the ten round game, which are shown in  Fig.~\ref{fig:withverticallines}. Five areas have been marked by using pairs of vertical dotted lines. In all five areas we can observe a small non binary adjustment in the first round penalty ($\ell^0_1$). The rightmost area is around $w=0.88$. More precisely, assume that the initial (first option) weight is $w^0_1=w=0.883$. Then, ten rounds are not enough to bring the weight under $1/2$; in fact a ten round penalty vector $(1,1, \ldots, 1)$ leaves the final (tenth) round weight at $w^9_1=0.5032$.
Therefore, a somewhat careless reading of Lemma~\ref{lemma:phaseonegameonly} would lead to the 
conclusion that the all one penalty vector is optimal.

However, a closer inspection of Fig.~\ref{fig:withverticallines} reveals that the optimal penalty of the first round is a little bit smaller than~1, the exact value being equal to $0.7968$. Indeed, the total cumulative loss for the penalty vector $(0.7968,1, \ldots, 1)$ is $7.1731$, while for $(1,1, \ldots, 1)$ it is $7.1704$. (The optimal total loss appears in Fig.~\ref{fig:numericalMultipleRoundParameterized4fig3}).
Actually, Lemma~\ref{lemma:phaseonegameonly} is  not applicable in this instance of the problem, since it requires the weight of an ``eleventh'' round to stay above $1/2$, which is not true (since $w^{10}_1=w \beta^{10}/(w \beta^{10}+1-w)=0.4476$).

An explanation of what has happened can be found in Table~\ref{table:unequalWeight10roundExample2data}. The loss in each round under the greedy policy is given in the first column, and the cumulative loss is given in the third column. The loss under the adjusted penalty vector is given in the third column, together with the cumulative loss in the last column. The adjustment in the first round creates a reduced loss, equal to $0.7273$, as compared to the maximum possible loss in this round, which equals $0.8830$. This initial sacrifice creates an opportunity for increased losses in the next nine rounds, as seen by comparing the first two columns. However, the total loss of the adjusted policy compensates the initial sacrifice for the first time in the last round.
An earlier compensation is not possible, since it would violate Lemma~\ref{lemma:phaseonegameonly}. Indeed  Lemma~\ref{lemma:phaseonegameonly} is still valid for  a nine round game, since $w^{9}_1 = 0.5032 > 1/2$, and this   guarantees the superiority of the greedy (binary) policy until the ninth round.
~$\Box$
\end{example}

\begin{table}
\centering
  \includegraphics[width=.3\textwidth]{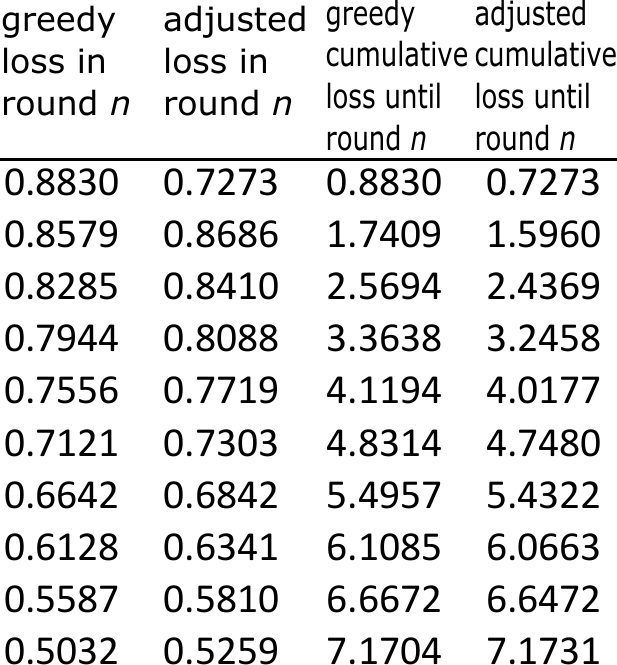}
  \caption{Loss in each round of the game of Example~\ref{example:microadjust}.}
  \label{table:unequalWeight10roundExample2data}
\end{table}

\begin{figure}
	\centering
  \includegraphics[width=.4\textwidth]{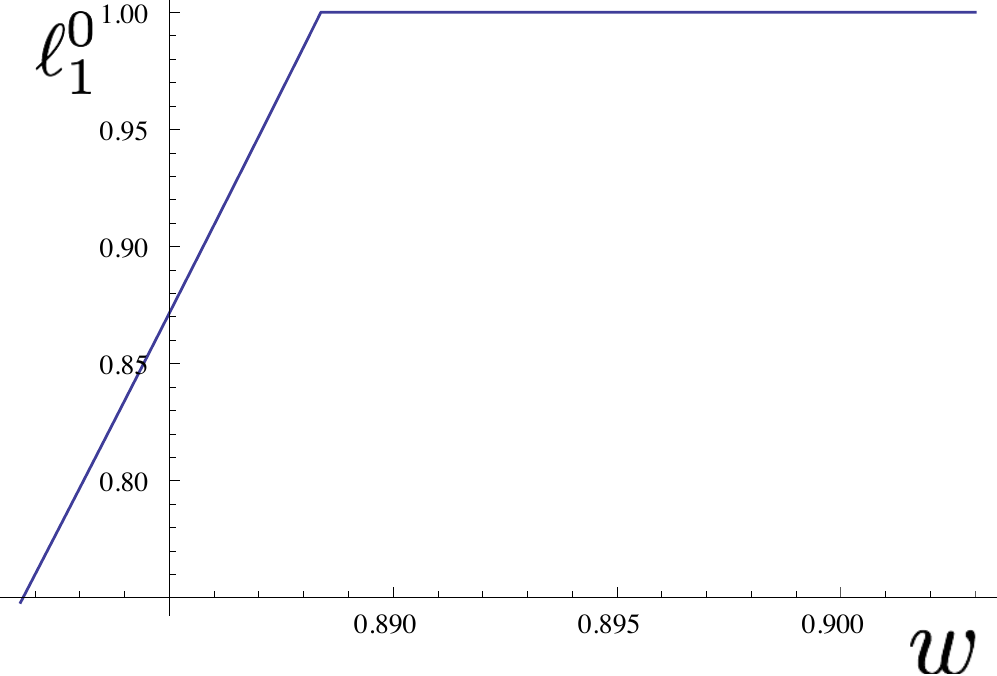}
  \caption{Optimal initial round first option penalty $\ell_1^0$ vs initial weight $w$ in the critical interval $\left[ \frac{1}{1 + \beta^{T-1}}, \frac{1}{1 + \beta^{T}} \right) = [0.882, 0.903)$ (for $\beta = 0.8$).}
  \label{fig:mixedcase4fig1}
\end{figure}

Let us explore somewhat further the above results. Lemma~\ref{lemma:phaseonegameonly} determines the optimal policy for game instances with initial weight $w$ in the interval  
$\left[\frac{1}{1 + \beta^{T}}, 1 \right]$,
while Lemma~\ref{lemma:marginal} determines game instances such that $w \in \left[ \frac{1}{1 + \beta^{T-1}}, \frac{1}{1 + \beta^{T}} \right)$. Both sets of instances are such that a greedy adversary cannot reverse the balance of weights between options, i.e. the second option weight can never climb above the first option weight under any policy, and all first option weights in all rounds, i.e. weights $w_1^0, w_1^1, \ldots, w_1^{T-1}$ are always greater than 1/2. However, in the games of Lemma~\ref{lemma:marginal} the ``residual'' weight after the end of the game, i.e. the weight $w_1^{T}$ that would enter a (non-existing) $T+1$'th round, falls below 1/2 and below $w_2^{T}$ under the greedy policy. In Fig.~\ref{fig:mixedcase4fig1} we use the game instance of the previous example and show the optimal first round penalty as a function of $w$ in the critical interval $\left[ \frac{1}{1 + \beta^{T-1}}, \frac{1}{1 + \beta^{T}} \right)$. Fig.~\ref{fig:mixedcase4fig1} could be seen as a magnification of part of Fig.~\ref{fig:withverticallines}. However, in Fig.~\ref{fig:mixedcase4fig1} we can clearly see that $\ell_1^0$ is greedy in the rightmost part of this interval, while in the leftmost part it grows from around 3/4 to 1 in a manner which seems to be linear. These results say, among other things, that the area of validity of Lemma~\ref{lemma:phaseonegameonly} should be extended beyond 
$1/(1 + \beta^{T})$ towards smaller $w$ values.

Given the analysis of the previous paragraph, Lemmas~\ref{lemma:phaseonegameonly}  and 
\ref{lemma:marginal} could be combined in a stronger lemma that would state that the left side of (\ref{ineq:dilemma2}), i.e.
\begin{equation}
F_n(\epsilon) \equiv ( 1- \epsilon ) f(0) + \epsilon \, [1-f(0)] + \sum_{k=1}^{n} f(k-2 \, \epsilon) 
\label{eq:fcapitalfunction}
\end{equation}
is maximized for $\epsilon =0$, if $w>w_f$, where $w_f$ is a value in the interval $\left[ \frac{1}{1 + \beta^{T-1}}, \frac{1}{1 + \beta^{T}} \right)$. Taking into account that 
\[
f(k - 2 \epsilon) = \frac{w \beta^{k - 2 \epsilon}}{1-w + w \beta^{k - 2 \epsilon}} 
\]
and also taking into account certain properties of the q-digamma function, $F(\epsilon)$ can be expressed as
\begin{equation}
F_n(\epsilon) = ( 1- \epsilon ) w + \epsilon \, [1-w] + \frac{\psi_{\beta} \left( 1- \frac{\ln \frac{w-1}{w \beta^{-2 \epsilon}}}{\ln \beta} \right)}{\ln \beta} + \frac{\psi_{\beta} \left( 1+n- \frac{\ln \frac{w-1}{w \beta^{-2 \epsilon}}}{\ln \beta} \right)}{\ln \beta}
\label{eq:fcapitalfunction2}
\end{equation}
where $\psi_q(z)$ is the q-digamma function, i.e. $\psi_q(z) = \Gamma'_q(z)/ \Gamma_q(z)$ and 
$\Gamma_q(z)$ is the q-Gamma function \cite{qgamma,krasniqi2010some}.

\subsection{Inside the phase of rotation}
While we shall see that all optimization related adjustments must be realized before entering rotation, their justification stems from the behavior of Hedge in the rotation phase.
Given a sufficient number of rounds, and an initial weight outside the rotation area,
at some round $t_0$ the current (first option) weight $w^{t_0-1}_1=w$ will enter the intersection area. Also, a game may start with a weight $w_1^0=w$ inside the intersection area. In both scenarios the future of the game is in the phase of rotation. In the second scenario there will be no transitional phase, nevertheless the adversary will try to optimize the rest of the game. In the first scenario, the current weight $w$ is probably an outcome of previous optimization, which has been implemented in the transitional phase, i.e. in the first $t_0$ rounds ($t \in [0, t_0 -1]$).
Thus one would expect that in this scenario $w$ is already optimized in view of some pre-determined (also optimal) transition phase policy. Of course it wouldn't hurt to try to optimize the rest of the game, but the expected outcome should be the same as in the originally optimized plan. With these thoughts in mind we take a weight $w$ in the intersection area as our starting point and try to optimize the adversary's policy for the rest of the game.

Once in the intersection area the adversary has the opportunity to implement the greedy penalty policy $(\ell^{t_0}_1, \ell^{t_0 +1}_1, \ldots ) = (1,0,1,0,\ldots)$, which (assuming $w>1/2$) will create a sequence of alternating weights $(w, w \beta /(w \beta + 1-w), w , w \beta /(w \beta + 1-w), w, \ldots )$. It will also create a loss equal to 
$w + (1-w) /(w \beta + 1-w)$ in each pair of rounds. The advantage of this policy is that the adversary creates a loss greater than 1/2 in every round. In addition, we have seen in section~\ref{sec:optimalperiodicweights} that the loss in each pair of rounds is optimized if $w=w_1^*=1/(1+ \sqrt{\beta})$. In some games the adversary will have already controlled the transitional phase (if such a phase exists) so as to achieve the optimal weight at the time of entering the intersection area, i.e. $w^{t_0-1}_1=w_1^*$.
In some other games the sacrifice necessary in the transitional phase in order to reach $w_1^*$ cannot be justified by the gains in the last part of the game, and the adversary opts for an entry weight other than  $w_1^*$.  Anyway, if $w^{t_0-1}_1 \neq w_1^*$, the adversary cannot move from $w^{t_0-1}_1 = w$ to $w_1^*$ at zero cost. An exception occurs if $w$ is exactly equal to 1/2, since under this assumption any adjustment can be achieved immediately at zero cost. We take up this particular, but very interesting, case in   section~\ref{sec:equalweights}.

We now summarize the basic thesis for the rotational phase: By stepwise using lemmas \ref{lem:rotation1} and \ref{lem:rotation2} it can be shown that any multiple step (i.e. round) scenario, which starts with a weight $w$ inside the intersection area, can be optimized to a scenario that has only one adjustment step in one of the first two rounds or no adjustment at all, depending on the remaining number of rounds, which should be able to absorb the adjustment cost. If an adjustment can be justified, it will take place in the first or the second round (of the remaining game), depending on the relative positioning between between $w$ and the target weight, as shown in the aforementioned lemmas.

Effectively, if $w^0_1=w$ is already in the intersection area, the adversary will use 
a penalty vector of the form $(1-x, 0, 1, 0, \ldots )$ or $(1,x,1,0, \ldots)$. Since the remaining total cumulative loss is a function of one variable, even direct optimization w.r.t. $x$ is possible. If there is room for the transitional phase, i.e. if $w \beta^{T} > 1-w$, the adversary would first opt for a penalty of the form $(1,1, \ldots, 1)$ until the weight enters the intersection area. Upon entering this area a non binary penalty adjustment would be needed, in the first or second round. However, any adjustments are ``pulled'' towards the beginning or the game. Therefore a general game will start with penalties $(1,x, 1, \ldots)$ or $(1-x,1,1,\ldots)$ and continue with ``1''s until the weight enters the intersection area, whereupon it will continue with alternating binary penalties. The optimum $x$ can be calculated even by using direct optimization, since there are only two candidate functions to optimize.

Consequences of the above are (a)~that all penalties beyond the second round are binary, and (b)~that the problem is solvable in polynomial time.

Another consequence is that it is now possible to estimate the distance between the optimum solution and the greedy one.

\subsection{Estimation of error induced by using binary penalties}
\label{sec:errorestimation}

\paragraph{Category~I: Games that include a clear transitional phase only}
If   $w \beta^{T} > 1-w$, the greedy policy is the optimal, and the ``error'' of using a greedy solution is zero.

\paragraph{Category~II: Games that include a transitional phase that ``touches'' the intersection area}
There is this somewhat ``singular'' case, in which $w \beta^{T} \leq 1-w < w \beta^{T-1}$, i.e. the assumptions of Lemma~\ref{lemma:marginal} are true. If the adversary adopts a greedy policy,  
the last round  still starts with a weight above 1/2, but  in an additional round (although an additional round does not exist) the weight would fall below 1/2.
We have seen that in such an instance of the problem a small initial sacrifice will optimize the total cumulative loss, by using a first round penalty equal to $1-\epsilon$. In mathematical terms this means that the maximum total cumulative loss can be expressed as $\max_{\epsilon} L(\epsilon)$, where
\[
L(\epsilon) = w (1- \epsilon ) + (1-w) \epsilon + \sum_{i=1}^{T-1} \frac{w \beta^{i-\epsilon}}{w \beta^{i-\epsilon} + (1-w) \beta^{\epsilon}}.
\]
Taking into account that the greedy policy loss is $L(0)$, we can calculate the ``binary penalty error'' as the difference 
\[ \max_{\epsilon} L(\epsilon) - L(0), \]
but we shall give a simple upper bound to this error.
For $w>1/2$ and $0< \epsilon <1$ the inequality $w (1- \epsilon ) + (1-w) \epsilon < w$ is true, therefore 
 \[
 L(\epsilon) - L(0) < \sum_{i=1}^{T-1} \frac{w \beta^{i-\epsilon}}{w \beta^{i-\epsilon} + (1-w) \beta^{\epsilon}} - \sum_{i=1}^{T-1} \frac{w \beta^{i}}{w \beta^{i} + 1-w }.
 \]
Note also that the summation terms grow as $\epsilon$ moves from 0 towards 1. If we accept the conjecture that the value of $\epsilon$ that maximizes $L(\epsilon)$ is in $[0,1/4]$, we can use $\epsilon = 1/4$ in the summation and obtain the following upper bound:
\[
L(\epsilon) - L(0) < \sum_{i=1}^{T-1} \frac{w \beta^{i-1/4}}{w \beta^{i-1/4} + (1-w) \beta^{1/4}} - \sum_{i=1}^{T-1} \frac{w \beta^{i}}{w \beta^{i} + 1-w }
\]
This inequality holds for any $\epsilon \in [0,1/4]$, thus it also holds for the $\epsilon$ that maximizes the total loss.

However, it is possible to obtain a looser, but much easier to calculate, upper bound by letting $\epsilon = 1/2$. This value of $\epsilon$ will further improve the summation, therefore
\begin{eqnarray*}
L(\epsilon) & < & w + \sum_{i=1}^{T-1} \frac{w \beta^{i-1/2}}{w \beta^{i-1/2} + (1-w) \beta^{1/2}} \\
& = & w +  \sum_{i=1}^{T-1} \frac{w \beta^{i-1}}{w \beta^{i-1} + (1-w) } \\
& = & w + \sum_{i=0}^{T-2} \frac{w \beta^{i}}{w \beta^{i} + (1-w) } \\
& = & w + L(0) - \frac{w \beta^{T-1}}{w \beta^{T-1} + 1-w } 
\end{eqnarray*}
and since the last round weight is still above 1/2 the following inequalities are also valid:
\begin{equation}
 L(\epsilon) - L(0)  < w - \frac{w \beta^{T-1}}{w \beta^{T-1} + 1-w } < w - \frac{1}{2} < 1/2
 \label{eq:category2bounds}
\end{equation}
Effectively, the new simplified, albeit very loose, bound is just $1/2$, and one interpretation of this result is that the error cannot me more than the loss of any single round (since in any round the loss is $\geq 1/2$). The true error is usually smaller by a few orders of magnitude. In Example~\ref{example:balancedgamecontinued} we present a 10 round game with $\beta=0.8$ and initial weight $w=0.883$. The maximum total loss is equal to $7.1731$ and occurs for $1-\epsilon = 0.7968$, while the greedy policy gives $7.1704$, therefore the error due to binary penalties is $0.0027$.

\begin{figure}
	\centering
  \includegraphics[width=.5\textwidth]{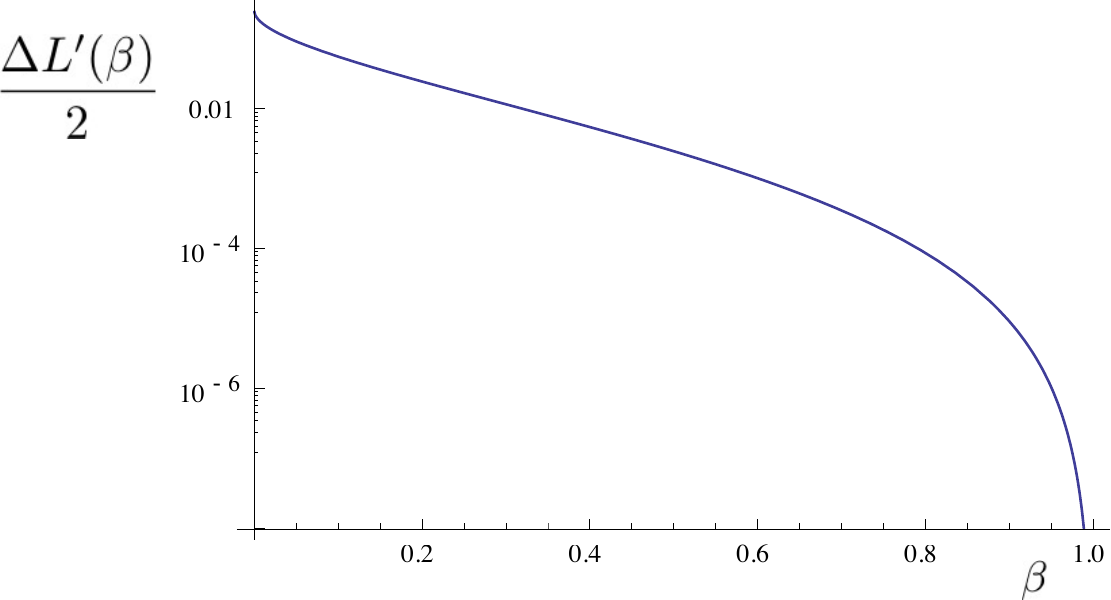}
  \caption{$\Delta L' (\beta)/2$, i.e. maximum error per round   due to binary policy, vs $\beta$.}
  \label{fig:rotationError1LogPlot}
\end{figure}

\paragraph{Category~III: Games that include rotation} The greedy policy eventually 
pushes the weights into the intersection area. If the $w'$ is the (first option) weight that appears for the first time in the intersection area (and always assuming that $w>1/2$), 
then the (first option weight) oscillates between $w'$ and $w' \beta /(w' \beta + 1-w')$, thereby producing a loss equal to
\[
L'=w' + 1 - \frac{w' \beta}{w' \beta + 1-w'}
\]
for each successive couple of rounds until the end of the game. $L'$ is maximized by the the ideal rotational weights of section~\ref{sec:optimalperiodicweights}, the maximum being equal to
$L_p=2/(1+ \sqrt{\beta})$. The exact value of the intersection area entry weight $w'$ is ``random'', as it totally depends on the initlal weight $w_1^0=w$ (and $\beta$).
On the other hand $L'$ is minimized for $w'=1/(1+\beta)$ or $w'=1/2$, which are also ``mirror'' values under rotation (i.e. $1/(1+\beta)$ with a first option penalty equal to~1 produces a weight equal to $1/2$ in the following round, and $1/2$ with a penalty equal to~0 produces $1/(1+\beta)$ in the following round). 
Both $w'=1/(1+\beta)$ and $w'=1/2$ give the minimum $L'$ as equal to $1/2 + 1/(1+\beta)$
Therefore the distance between the maximum and the minimum value of $L'$ is
\[
\Delta L' (\beta)= \frac{2}{1+ \sqrt{\beta}} - \frac{1}{2} - \frac{1}{1+\beta} \geq 0
\]
which expresses the maximum error in loss due to binary penalties per couple of rounds. Note that $\Delta L' (0) = 1/2$, $\Delta L' (1) = 0$, and $0 \leq \Delta L' (\beta)\leq 1/2$. Fig.~\ref{fig:rotationError1LogPlot} gives the average error per round $\Delta L' (\beta)/2$ vs $\beta$. Although for $\beta$ close to zero the per round error can be as high as 1/4, for $\beta$ greater then 1/2 it falls rapidly below 1/1000, and for $\beta = 0.8$ the error is smaller than $10^{-4}$.

Let us summarize our findings. Assuming a game such that $w \beta^{T} < 1-w$, i.e. a game such that there exists at least one round in the intersection area, the game in the first $T_1$ rounds falls in the previous Category~II, where
\begin{equation} 
w \beta^{T_1} \leq 1-w <  w \beta^{T_1 -1} \Leftrightarrow T_1 = \left\lceil \frac{\ln (1-w) - \ln w}{\ln \beta} \right\rceil 
\label{eq:transitionphaselength}
\end{equation}
As we have already seen, in Category~II games the optimal policy for the adversary is either simply binary, or there is a small adjustment in the first or the second round, and for the error one can use any of the upper bounds given by (\ref{eq:category2bounds}). In fact these bounds have been calculated for a game of duration $T_1$ exactly. However, here we are exploring a longer game of $T >T_1$ rounds, which means that the total cumulative loss for the $T$ round game may be suboptimal w.r.t. to a $T_1$ round game, since in the $T$ round game the adversary has to cater for the remaining $T-T_1$ rounds as well. Effectively, the bounds given by 
(\ref{eq:category2bounds}) remain valid for the $T$ round game, but one could try to find a tighter bound.

Since the next $T-T_1$ rounds are rotational rounds with an error upper bounded by $\Delta L' (\beta)$ in each pair of rounds, the total error in the last $T-T_1$ rounds is bounded by
\begin{equation}
 \left\lceil \frac{T-T_1}{2} \right\rceil \Delta L' (\beta)
\label{eq:boundofrotationalphase}
\end{equation}
This expression actually contains the error of an additional (non existing) round if the rotation contains an even number of rounds.

Finally, the total error can be calculated as the sum of the bound of the transitional phase (\ref{eq:category2bounds}) for the first $T_1$ rounds (where $T_1$ is given by \ref{eq:transitionphaselength}), and the bound of the rotational phase given by (\ref{eq:boundofrotationalphase}).

Practically speaking, the transitional phase error does rarely exceed the order of magnitude of $10^{-3}$, and the per round error in the rotational phase is usually below $10^{-4}$ for reasonably valued $\beta$. Therefore, the total error of the binary policy will accumulate to a figure that is comparable to the loss of a single round only in really long games with thousands of rounds in the rotational phase.

\section{The special case of equal initial weights}
\label{sec:equalweights}
The natural choice for a player that is not in possession of additional or internal information is to start the game with equal initial weights, i.e. $w=w_1=w_2=1/2$ (if $N=2$). 
Moreover, we have seen in all results that have so far been presented that the maximum loss achievable by an intelligent adversary is minimized for $w=1/2$. Therefore it is in the players best interest to start with $w=1/2$.

On the other hand games that start with equal weights are easier to analyze: The sacrifice that is often necessary for optimally entering rotation comes ``for free''  if the weights are equal, and, fortunately, this fact contributes to a closed form solution, as we shall show in this section.
In \cite{anagnostou2015worst} we have given a short duration game example ($T=3$) in which the initial weights are $(1/2,1/2)$. The overall maximization of the total loss has produced a pair of penalties $(3/4,1/4)$ for the first round and the pairs $(0,1)$, $(1,0)$ for the remaining two rounds.
This is a remarkable result in the light of the analysis of this section, since in just one round the adversary ``converts'' the initial weights $(1/2,1/2)$ to the ideal pair of weights (given by (\ref{idealpairofweights}) in section~\ref{sec:optimalperiodicweights}) and is able to use them 
in the next two rounds. In this particular example the initial ``adjustment'' is made possible without any sacrifice for the adversary, since any pair of penalties ($\ell, 1- \ell$) produce a constant first round loss (equal to $\frac{1}{2} \times \ell + \frac{1}{2}(1- \ell)= \frac{1}{2}$) for any $\ell$.

Moreover, Example~\ref{exmpl:equalweights} provides a strong hint for  the analysis of instances of Problem~\ref{problem:general1} that start with equal weights and extend their duration for an arbitrary number of rounds $T$. 
For $N=2$ the obviously suggested solution is that the adversary should use the penalty sequence
\[
(\ell_1^0, \ell_1^1, \ell_1^2, ..., \ell_1^{T-1}) =
\left(\frac{3}{4}, 0 , 1 , 0 , 1 , \ldots \right)
\]
or the sequence
\[
(\ell_1^0, \ell_1^1, \ell_1^2, ..., \ell_1^{T-1}) =
\left( \frac{1}{4}, 1, 0 , 1 , 0 ,  \ldots \right)
\]
Both schemes will produce a total loss equal to $ 1/2+ {T-1}/{(1+ \sqrt{\beta})}$,
which can be seen as a lower bound of the total cumulative loss for equal initial weights, i.e.
\begin{equation}
L_{\max}^{T-1} \left( \frac{1}{2} \right) \geq  \frac{1}{2}+ \frac{T-1}{1+ \sqrt{\beta}},
\label{eq:lowerboundequalweights}
\end{equation}
 but we already know that this is the correct maximum loss for a three round game, i.e. $L_{\max}^{2}(1/2) =  1/2+ 2/(1+ \sqrt{\beta})$. Therefore, a reasonable conjecture to be explored is whether the equality would still be valid for any $T$, since the weight adjustment is in accordance with Lemma~\ref{lemma:targetweightaboveinters}, as it takes place in the very first round.
Numerical tests based on the methodology of section~\ref{sect:recursion} reveal  that (\ref{eq:lowerboundequalweights}) holds as an equality for odd values of $T$, but it is a strict inequality for even values. Let us see why.

Suppose that the adversary uses the penalty sequence $(\ell_1^0, \ell_1^1, \ell_1^2, ..., \ell_1^{T-1}) =
\left(x, 0 , 1 , 0 , 1 , \ldots \right)$. Since the initial weights are the initial weights $(w_1^0, w_2^0) = (1/2,1/2)$, the weights in the second round are 
\[
w_1^1 = \frac{\beta^x}{\beta^x + \beta^{1-x}},  \quad w_2^1 = \frac{\beta^{1-x}}{\beta^x + \beta^{1-x}}
\] 
and the weights in the third round are
\[
w_1^2 = \frac{\beta^{x}}{\beta^x + \beta^{2-x}},  \quad w_2^2 = \frac{\beta^{2-x}}{\beta^x + \beta^{2-x}}
\] 
The total loss in both the second and third round (i.e. the cumulative loss in one cycle) is
\[
L_c = \ell^1_1 w_1^1 + \ell^1_2 w_2^1+ \ell^2_1 w_1^2 + \ell^2_2 w_2^2 = w_2^1 + w_1^2 = 
\frac{\beta^{1-x}}{\beta^x + \beta^{1-x}} + \frac{\beta^{x}}{\beta^x + \beta^{2-x}}
\]
If $T=2k+1$, the total cumulative loss achieved by using the aforementioned penalty scheme is
\[
L_p = \frac{1}{2} + kw_2^1 + k w_1^2  = \frac{1}{2} + k \frac{\beta^{1-x}}{\beta^x + \beta^{1-x}} + k \frac{\beta^{x}}{\beta^x + \beta^{2-x}},
\]
which is maximized for $x=3/4$ as seen before. This value of $x$ produces equal terms $w_2^1$ and $w_1^2$.
If, however, $T=2k$, then
\[
L_p  = \frac{1}{2} + kw_2^1 + (k-1) w_1^2 = \frac{1}{2} + k \frac{\beta^{1-x}}{\beta^x + \beta^{1-x}} + (k-1) \frac{\beta^{x}}{\beta^x + \beta^{2-x}},
\]
since the last cycle is not complete.
In the latter case the adversary can take advantage of the fact that the term  $w_2^1$  appears more times than the term  $w_1^2$ and  increase the first term at the expense of the second  by using a value of $x$ higher than $3/4$. The smaller the number of rounds $T$, the more will the adversary try to increase $w_2^1$  w.r.t.  $w_1^2$, but of course $x$ cannot exceed~1.

\begin{figure}
	\centering
  \includegraphics[width=.75\textwidth]{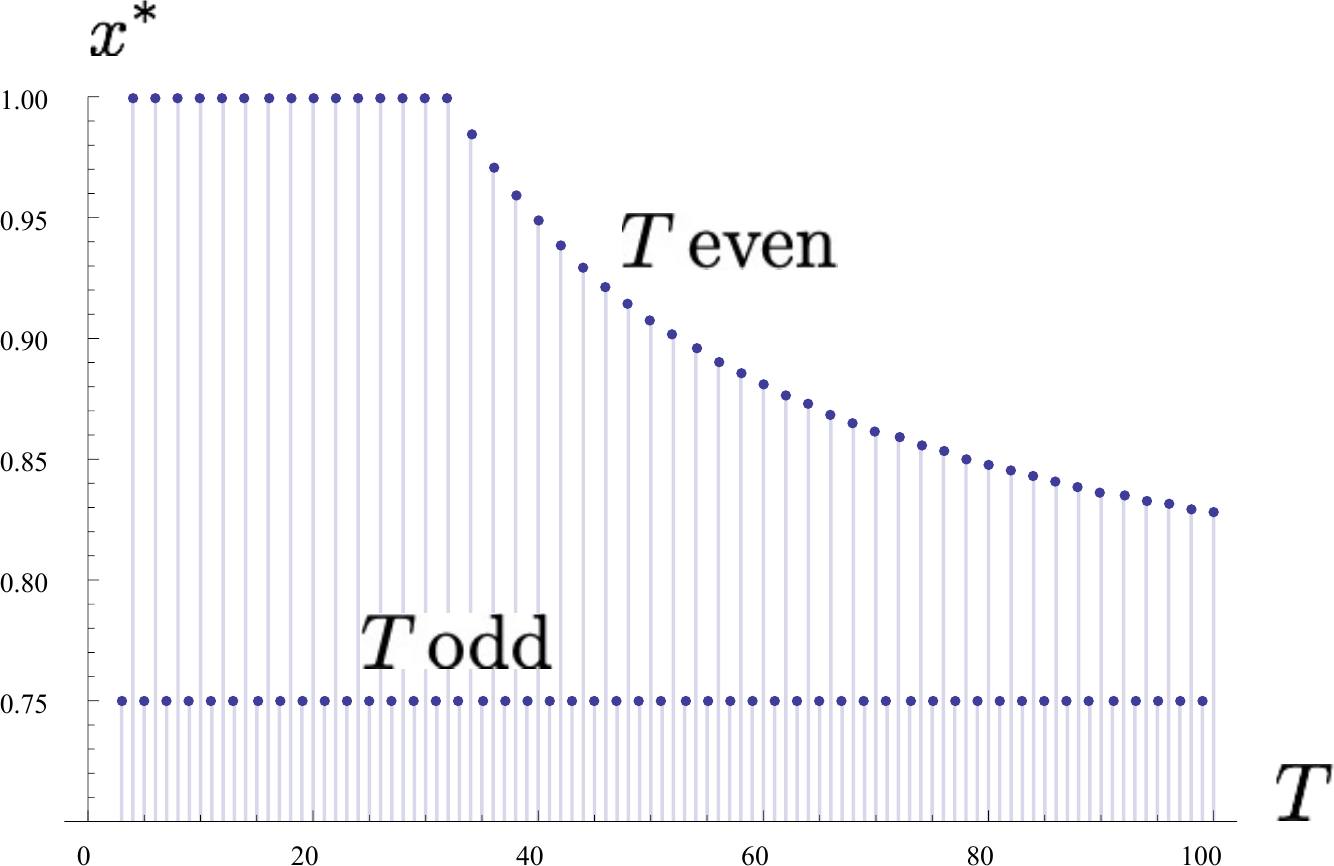}
  \caption{Optimal first round penalty $x^*$ vs $T$, for $\beta = 0.6$.}
  \label{fig:equalweightevenTmaxi2a}
\end{figure}

The above two expressions can be unified in the following expression that is valid for both odd and even values of $T$:
\begin{equation}
L_p = \frac{1}{2} + \left\lceil \frac{T-1}{2} \right\rceil \frac{\beta^{1-x}}{\beta^x + \beta^{1-x}} + \left\lfloor \frac{T-1}{2} \right\rfloor \frac{\beta^{x}}{\beta^x + \beta^{2-x}}
\label{eq:evenTLp}
\end{equation}
In Fig.~\ref{fig:equalweightevenTmaxi2a} we present a graph of the optimal first round penalty $x^*$ (that maximizes the total cumulative loss $L_p$ of formula (\ref{eq:evenTLp})) vs $T$ for $\beta = 0.6$.
The figure shows that for games with even $T$
no adjustment is necessary in the initial round if  $T \leq 32$; thus the adversary can adopt a totally binary (and greedy) policy. If $T$ is still even, but greater than 32, $x^*$ becomes fractional and tends asymptotically to the optimal value for games with an odd number of rounds, i.e. to $3/4$. Of course, if $T$ is odd, then $x^*$ is always equal to $3/4$.
\begin{figure}
	\centering
  \includegraphics[width=.5\textwidth]{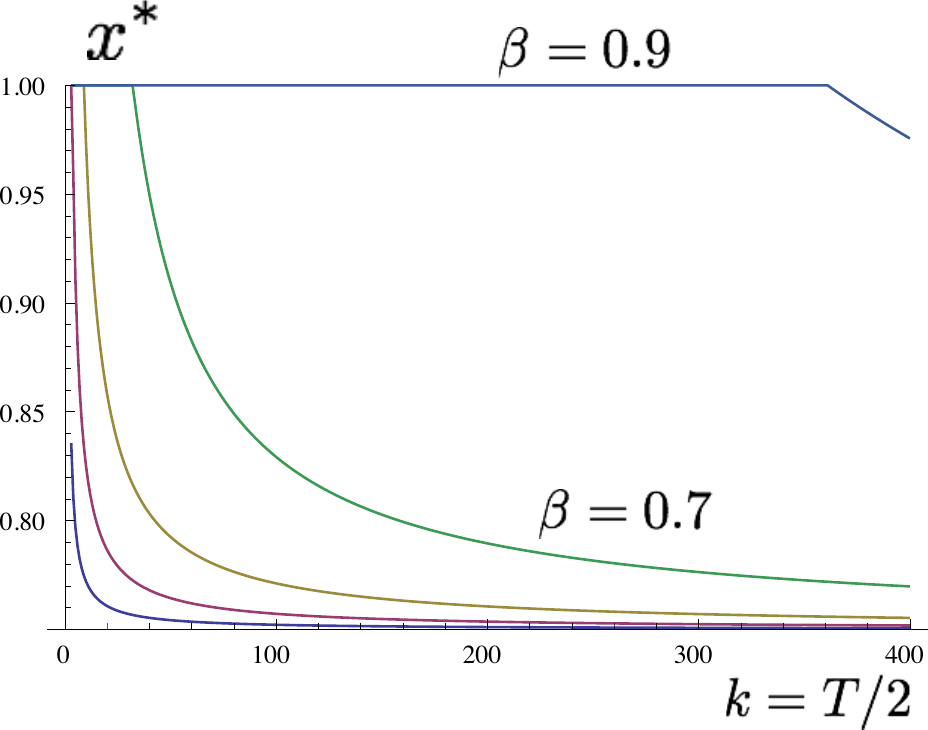}
  \caption{Optimal first round penalty $x^*$ vs $k=T/2$, for $\beta = 0.1,0.3,0.5,0.7,0.9$.}
  \label{fig:equalweightevenTmaxi1a}
\end{figure}

Clearly, if $T$ is odd, the adversary will be able to maximize the player's loss by using penalties $(3/4,1/4)$ (or vice versa) in the first round, as this action would immediately generate the optimal pair of weights given by (\ref{idealpairofweights}). In the next $T-1$ rounds the greedy (binary) policy is optimal. If $T$ is even, $x^*$ is between $3/4$ and~1. An adversary who is ignorant of this kind of fine tuning and uses a greedy policy (with perhaps a  random choice in the first round, since the initial weights are equal), will behave suboptimally in both cases. In the first case (when $T$ is odd), there will be a constant loss in each round. In the second case  (when $T$ is even), the required adjustment is smaller, therefore the lack of adjustment implies a less pronounced suboptimality, and under certain conditions ($\beta$ large enough, $T$ small enough) $x^*=1$, i.e. there is no suboptimality at all. A different kind of ``ignorance'' is involved if the adversary uses the rotationally optimal pair (by always setting $x^*=3/4$) without taking into account the parity of $T$. Then the relative error is more pronounced in shorter games than in longer ones.  However, in most of these situations the suboptimality is practically insignificant.
For example, if $\beta=0.7$, the relative error is less than $1\%$ in each round. 

In conclusion, the expression~\ref{eq:lowerboundequalweights} for the total cumulative loss of two option games that start with equal weights can now be replaced by the more accurate expression
\begin{equation}
L_{\max}^{T-1} \left( \frac{1}{2} \right) = \frac{1}{2} + \left\lceil \frac{T-1}{2} \right\rceil \frac{\beta^{1-x^*}}{\beta^{x^*} + \beta^{1-x^*}} + \left\lfloor \frac{T-1}{2} \right\rfloor \frac{\beta^{x^*}}{\beta^{x^*} + \beta^{2-x^*}},
\label{eq:equalweights1}
\end{equation}
However,  for odd valued $T$ we have seen that $x^*=3/4$, and (\ref{eq:equalweights1})  simplifies to
\begin{equation}
L_{\max}^{T-1} \left( \frac{1}{2} \right) = \frac{1}{2}+ \frac{T-1}{1+ \sqrt{\beta}},
\label{eq:equalweights2}
\end{equation}
Also, for $T=2k$ it is possible to achieve a closed form solution for $x^*$. The first step towards the evaluation of $x^*$ is to replace $b^x$ by a new variable, say $y$, in (\ref{eq:evenTLp}). Then by  setting  $\partial L_p / \partial y =0$, choosing the right (positive) root (out of five possible roots), going back to $x$, and accepting a value up to~1, one will obtain the following result:
\begin{equation}
x^* =\min \left[1, \frac{ \ln \left( - \frac{\beta^2 - \sqrt{(\beta-1)^2 \beta^3 k (k-1)}}{\beta+k-\beta k} \right) }{\ln \beta} \right]
\label{eq:xstar}
\end{equation}
There is also a symmetric solution with the same total loss obtained by using the reverse penalties $\left( 1-x^*, 1, 0 , 1 , 0 ,  \ldots \right)$. 
Expressions (\ref{eq:equalweights1}), (\ref{eq:equalweights2}), and (\ref{eq:xstar}) give a complete closed form solution for the significant category of games that start with equal weights between options.

\section{Further examples}
Perhaps the best way to illustrate the optimal behavior of the adversary is to revisit the sample problem that has already been solved by using the recursive methodology of section~\ref{sect:recursion}.

\begin{figure}
	\centering
  \includegraphics[width=.6\textwidth]{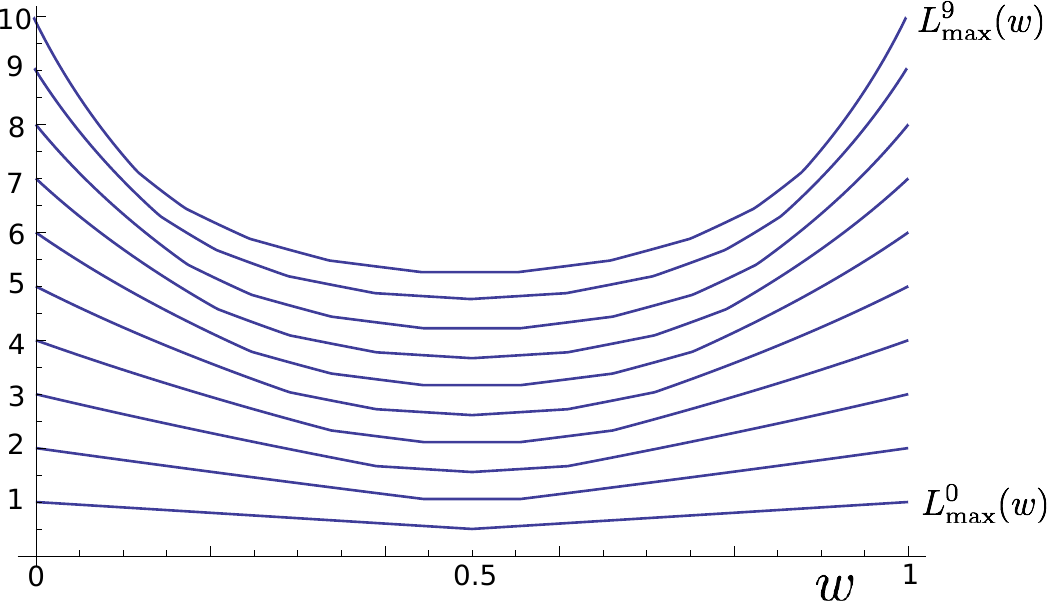}
  \caption{Total cumulative loss vs $w$ for $T=1,2,\ldots,10$, and for $\beta=0.8$.}
  \label{fig:numericalMultipleRoundParameterized6afig3}
\end{figure}

\begin{example}
\label{example:balancedgamecontinued}
We revisit Example~\ref{example:balancedgame} in the light of the previous analysis.
We can easily calculate the number of rounds, in which Lemma~\ref{lemma:phaseonegameonly} remains valid, i.e. the current (first option) weight remains above $1/2$ and cannot fall below $1/2$ in one more round. 
Assuming that the game is long enough, the length of this phase (measured in rounds) is equal to  $t_1$, where  $t_1$  is such that
\[
w \beta^{t_1+1} \geq1-w \Leftrightarrow t_1 \leq \frac{\ln \left( \frac{1-w}{w} \right)}{\ln \beta} -1,
\]
and
\[
w \beta^{t_1+2}<1-w  \Leftrightarrow t_1 > \frac{\ln \left( \frac{1-w}{w} \right)}{\ln \beta} -2,
\]
If the game is not long enough, i.e. of $T< t_1$, then of course the duration of this first phase will be equal to $T$.
There will be no first phase at all if the starting weight $w$ is  too close to 1/2. Therefore the first phase duration $T_1$ is given by
\begin{equation}
T_1 = \max \left\{ \min  \left\{ \left\lfloor \frac{\ln \left( \frac{1-w}{w} \right) }{\ln \beta} -1 \right\rfloor , T \right\} ,0 \right\}
\label{eq:T1}
\end{equation}
A plot of $T_1$ vs $w$ is shown in Fig.~\ref{fig:firstPhaseDuration} for the purpose of comparison with the penalties of Fig.~\ref{fig:withverticallines}.  
For $w>0.92$  Lemma~\ref{lemma:phaseonegameonly} is valid, there is no second phase at all, and the  adversary can fully adopt the greedy approach, i.e.  ten successive unitary penalties for option~1. 

\begin{figure}
	\centering
  \includegraphics[width=.4\textwidth]{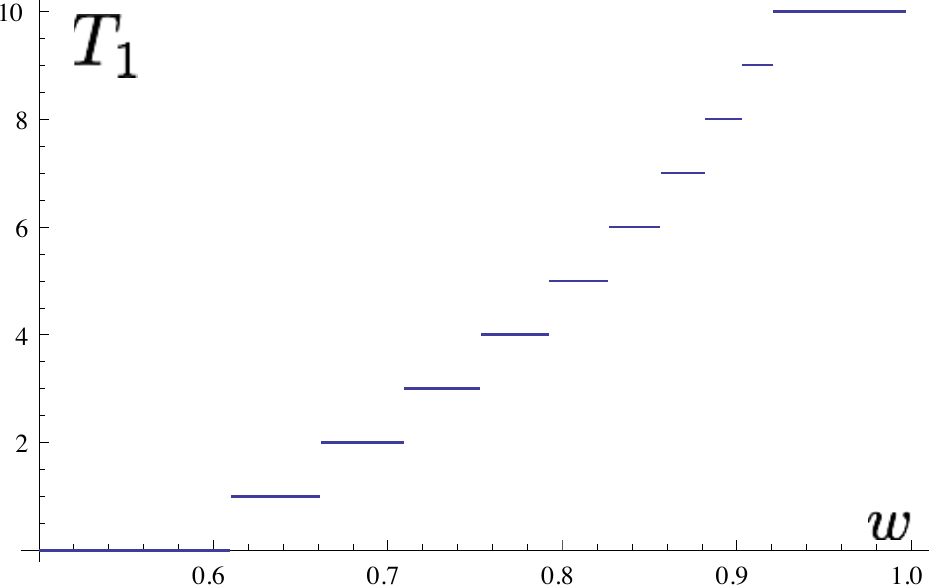}
  \caption{First phase duration $T_1$ vs initial weight $w$ (for $w>1/2$ and $\beta = 0.8$).}
  \label{fig:firstPhaseDuration}
\end{figure}

As $w$ decreases, at $w$ around $0.9$ we reach an area, which according to Fig.~\ref{fig:numericalMultipleRoundParameterized4fig2} makes the adversary choose a non binary first round penalty. 
For example, consider the weight $w=0.883$. Then the assumptions of Lemma~\ref{lemma:marginal} become true and the greedy penalty vector
\[
(\ell^0_1 \ldots \ell^9_1) = (1,1,1,1,1,1,1,1,1,1),
\]
which produces a total cumulative loss equal to $7.1704$, is no longer optimal.
Nevertheless, the weight in the final round is equal to $w^9_1=0.5032$, i.e. still above 1/2, therefore a greedy adversary would still choose $p^9_1=1$, which is possibly suboptimal.

In order to make use of Lemma~\ref{lemma:marginal}  let us set 
$\ell^0_1 = 1-\epsilon$ and $w_1^0=w$. Then the total loss to be maximized w.r.t. $\epsilon$ is equal to
\[
( 1-\epsilon) w + \epsilon (1-w) + \sum_{i=0}^{T-2} \frac{w \beta^{ 1-\epsilon+i}}{w \beta^{ 1-\epsilon+i}+ (1-w) \beta^{ \epsilon}}
\]
Direct optimization w.r.t. $\epsilon$ (for $w=0.883$) gives the optimal penalty vector as
\begin{equation}
(\ell^0_1 \ldots \ell^9_1) = (0.7968,1,1,1,1,1,1,1,1,1),
\label{eq:adjustbeforetenrounds}
\end{equation}
which results in a total loss equal to $7.1731$. 

\begin{figure}
	\centering
  \includegraphics[width=.4\textwidth]{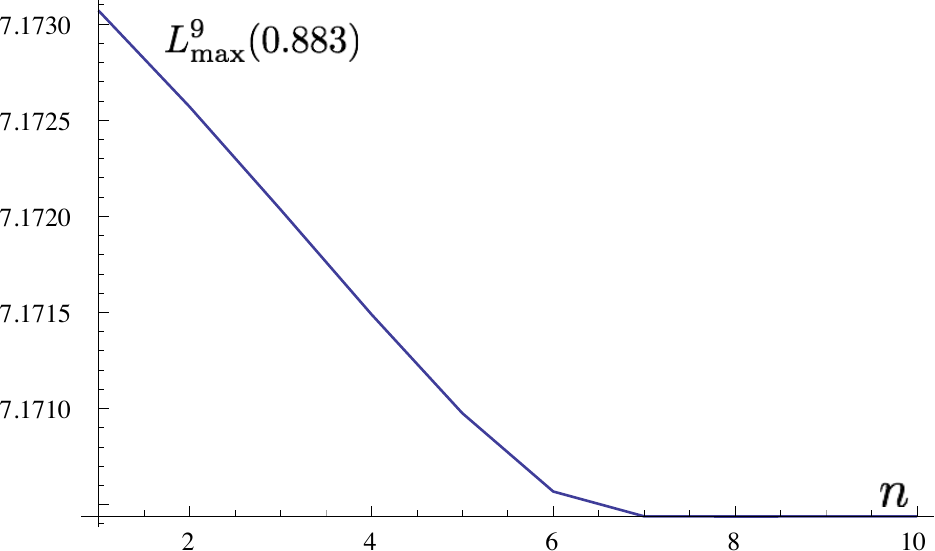}
  \caption{Total cumulative loss if a non binary adjustment is optimally executed in the $n$'th round ($n=1,2,\ldots,10$).}
  \label{fig:explicit10roundMaximization2fig2}
\end{figure}

According to Lemma~\ref{lemma:marginal} any necessary adjustments should  be undertaken as soon as possible. Fig.~\ref{fig:explicit10roundMaximization2fig2} shows the effect of pushing the adjustment beyond the initial round on the total cumulative loss.

Further examination of Fig.~\ref{fig:withverticallines} shows that for $w=0.556$ there is one more non binary adjustment in the first round, but in the neighboring $w=0.555$  the adjustment moves into the second round; this calls for an explanation.
It seems that there is a kind of micro-management at this point.
If $T$ were very large, the adversary would try to approach the optimal rotation weights $(1/(1+\sqrt{\beta}),\sqrt{\beta}/(1+\sqrt{\beta})$ as soon as possible. Since $w> 1/2$, the adversary would take an adjustment step from $w$ to $1/(1+\sqrt{\beta})$ (because $\sqrt{\beta}/(1+\sqrt{\beta})$ is not reachable in one step).

Finally, we give a graph of the total cumulative loss vs $w$ for $T=1,2,\ldots,10$ in Fig~\ref{fig:numericalMultipleRoundParameterized6afig3}.
The top curve is the same as in Fig.~\ref{fig:numericalMultipleRoundParameterized4fig3}.
Note, however, that the penalties for each curve must be calculated separately. In other words, the penalties that create the loss of the ninth curve are {\em not} the first nine penalties of Fig.~\ref{fig:withverticallines}.~$\Box$ 
\end{example}

The ten round game of Example~\ref{example:balancedgame} is balanced in the sense that the Hedge adaptation factor $\beta$ is 
not excessively strong and each phase consists of several rounds if 
$w$ is not very close to the values $0,1/2,1$. In the next example we explore games with an exceptionally  strong $\beta$ factor.

\begin{example}
\label{example:strongbeta}
We set $\beta = 0.1$ and $T=10$ rounds again.
The quantization of $w$ is as before.
The resulting total cumulative loss vs $w$ is shown on Fig.~\ref{fig:numericalParameterized7beta01fig1}, while Fig.~\ref{fig:numericalParameterized7beta01fig2} shows the associated penalties. In order for the first phase to last even for one round a value $w$ that satisfies $w \beta > 1-w$ is needed, which gives $w > 1/(1+\beta) = 10/11 \simeq 0.9091$.~$\Box$

\begin{figure}
	\centering
  \includegraphics[width=.6\textwidth]{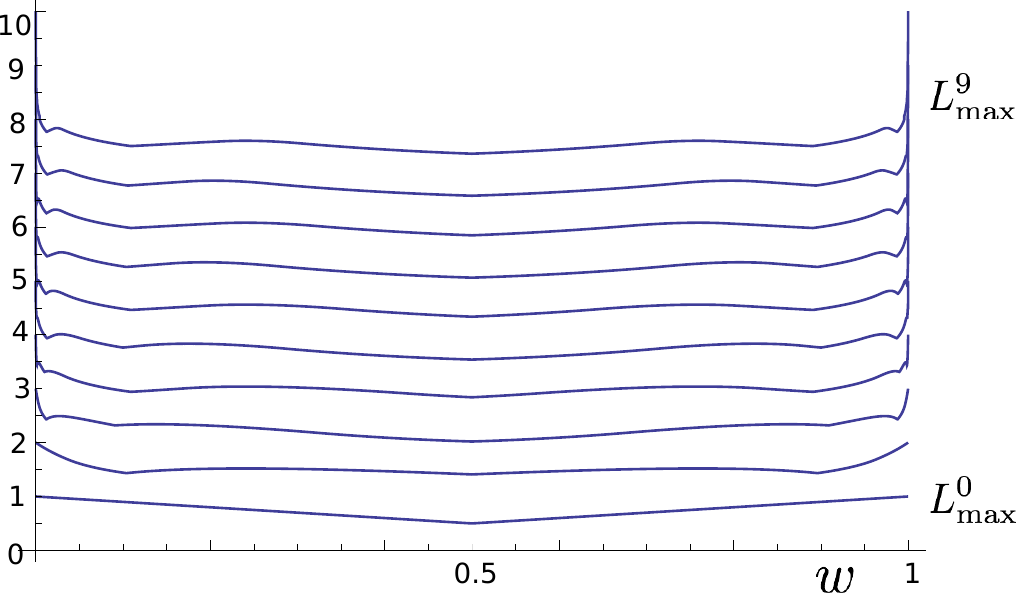}
  \caption{Total cumulative loss after $T=10$ rounds vs initial (first option) weight $w$  (for $\beta = 0.1$).}
  \label{fig:numericalParameterized7beta01fig1}
\end{figure}

\begin{figure}
	\centering
  \includegraphics[width=.6\textwidth]{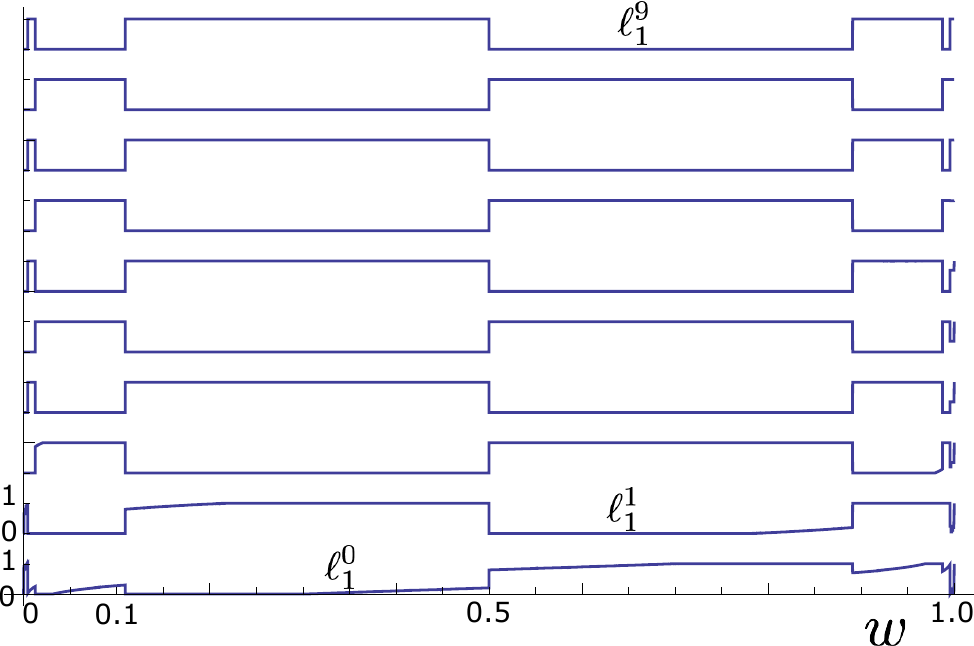}
  \caption{Round $i$ first option penalty $\ell_1^i$ ($i=0,1,\ldots,9$) vs initial (first option) weight $w$ (for $\beta = 0.1$).}
  \label{fig:numericalParameterized7beta01fig2}
\end{figure}

\end{example}


Assume that a game is long enough, so as to enable the adversary to reach rotation, i.e. $T$ is such that $w \beta^{T} < 1-w$. Let $T_0$ be the minimum number of rounds that enables the adversary to enforce rotation by using greedy binary penalties, i.e. $T_0$ is such that $w \beta^{T_0} < 1-w$. Effectively, the adversary can force Hedge to enter rotation  after $T_0$ rounds for the rest of the game, i.e. for the next $T-T_0$ rounds, by setting
\[
\left( \ell^{T_0}, \ell^{T_0+1} , \ldots, \ell^{T-1} \right) = (0,1, \ldots)
\]
This sequence creates a rotation that involves cycles consististing of two rounds each with the pair of weights
\begin{equation}
\left(\frac{w \beta^{T_0}}{w \beta^{T_0}+1-w}, \frac{1-w}{w \beta^{T_0}+1-w} \right)
\label{eq:pair1}
\end{equation}
in the first round and the pair
\begin{equation}
\left(\frac{w \beta^{T_0 -1}}{w \beta^{T_0 -1}+1-w}, \frac{1-w}{w \beta^{T_0 -1}+1-w} \right)
\label{eq:pair2}
\end{equation}
in the second round. This scheme produces a loss per cycle equal to
\begin{equation}
L_c= \frac{1-w}{w \beta^{T_0}+1-w} + \frac{w \beta^{T_0 -1}}{w \beta^{T_0 -1}+1-w}
\label{eq:cycleloss}
\end{equation}
According to the analysis of section~\ref{sec:optimalperiodicweights}, in this period the adversary could
have achieved a per cycle loss equal to $2/ (1+\sqrt{\beta})$, which is greater than $L_c$ of formula~\ref{eq:cycleloss}. However, the migration from weights \ref{eq:pair1} and \ref{eq:pair2} to the ``ideal'' pair $\left( 1/ (1+\sqrt{\beta}), \sqrt{\beta}/ (1+\sqrt{\beta}) \right)$ (or vice versa) requires the aforementioned sacrifice, which involves less profitable fractional penalties in the transition step. If the remaining number of rounds $T-T_0$ is small, the gains after the transition may not justify the cost of the transition.

\section{The optimal adversarial algorithm}
In this final section we summarize the methodology that should be used by the adversary. However, let us first remind that the quick and dirty solution, which is subject to a very small error as explained in section~\ref{sec:errorestimation}, is  the (straightforward) greedy policy. A second observation is that the optimal policy that has been analyzed in this paper is clearly of polynomial complexity. Also, the optimal policy is not only non greedy, but any necessary adjustments, which in some sense may depend on how the game ends, are reflected in measures that must be taken in the beginning of the game. For example, the addition of one more round to a game may convert a greedy optimal policy to a non greedy optimal policy with an adjustment in the very first round. 

First, assume that an instance of a game is given with specific values for the number of rounds $T$, the adjustment parameter $\beta$, and the the initial weights $(w_1^0, w_2^0) = (w,1-w)$. Assume also that without loss of generality $w \geq 1/2$. If $w=1/2$ the approach of the adversary is as described in section \ref{sec:equalweights}. Therefore, in the following we assume that $w>1/2$. 

We give now a sketch of the main ideas that exist behind the algorithm that optimizes the adversary's behavior in a two option game against a player that uses Hedge. Let us use the greedy solution as a starting point.
The greedy adversary would  constantly penalize the first option until at some round the current (first option) weight falls below 1/2. Then a sequence of alternating zeros and ones $(0,1,0,1, \ldots)$ is used until the end of the game. Effectively the rotating weights that appear in the second phase depend on the initial weight $w$. The non greedy adversary follows the same basic approach, albeit with certain improvements:
\begin{enumerate}[label=\Alph*]
\item If the game is not long enough so as to make it into the rotation phase, the optimal policy is greedy.
\item If the game marginally misses the rotation phase, as described previously, an adjustment in the initial step may by beneficial for the adversary.
\item If the game is long enough so as to include a rotation phase, the adversary may try to approach the ideal rotating weights, even partially, by using a single non binary penalty somewhere in the game.
\item If the number of rounds in the rotation phase is odd, one of the two rotating weights appears once more than the other weight, and this should be taken into account in determining the target rotating weights.
\item Any non binary adjustment that aims at improving the rotating weights should be undertaken as early as possible, i.e. at the beginning of rotation or earlier. Effectively the rotating weights remain the same for the duration of the rotational phase.
\item The adjustment is done either by decreasing a penalty equal to~1 or by increasing a penalty equal to~0, depending on which choice involves a smaller sacrifice. If the adjustment involves decreasing an~1, it is should better be done in the beginning of the game. If the adjustment involves increasing a~0, it effectively appears as a correction to the first zero, which cannot appear before the beginning of the rotation phase. 
\item A small adjustment in the first round may occasionally also improve the performance of the transitional phase, but the final decision on this first penalty depends on the existence of a  rotating phase, and the target rotating weights.
\end{enumerate}

Effectively there is a limited number of optimal penalty patterns for the adversary as follows:
\begin{enumerate}
\item If $w, \beta$ and $T$ are such that $w \beta^{T} > 1-w$, the optimal penalty scheme is greedy, i.e. 
\[
\left( \ell^0_1, \ell^1_1, \ldots \ell^{T-1}_1 \right) = (1,1,\ldots,1)
\]
\item \label{step2} If $w \beta^{T-1} \geq 1-w$, and $w \beta^{T} <1-w$, i.e. $w \in \left[ \frac{1}{1 + \beta^{T-1}}, \frac{1}{1 + \beta^{T}} \right)$,
the adversary  can try
\[
\left( \ell^0_1, \ell^1_1, \ldots \ell^{T-1}_1 \right) = (1-\delta,1,\ldots,1),
\]
where $0 \leq  \delta <1$, 
and optimize the total cumulative weight as a single variable function of~$\delta$. The result of this optimization is usually a very small $\delta$ or exactly zero. A more detailed analysis shows that $\delta$ is an increasing function of $w$ in the interval $ \left[ \frac{1}{1 + \beta^{T-1}}, \frac{1}{1 + \beta^{T}} \right)$, and remains equal to~1 for $w>w'$, where $w'$ is a value inside this interval. 
\item \label{lab:3} If $w \beta^{T-1} < 1/2$, the adversary calculates the length of the transition phase $T_1$ by using (\ref{eq:T1}) and sets either
\begin{enumerate}[label=\alph*]
\item \[
\left( \ell^0_1, \ell^1_1, \ldots \ell^{T-1}_1 \right) = (\underbrace{1-\delta,1,1,\ldots,1}_{T_1},0,1,0,1,\ldots )
\]
if  the first weight that would appears in the intersection area under a greedy scheme would be below the ideal weight and above $1/2$,
or
\item \label{lab:3b}
\[
\left( \ell^0_1, \ell^1_1, \ldots \ell^{T-1}_1 \right) = (\underbrace{1,1,1,1,\ldots,1}_{T_1},\delta,1,0,1,\ldots )
\]
if the first weight that would appears in the intersection area under a greedy scheme would be above the ideal weight.
\end{enumerate}
In both cases the adversary can optimize a single valued function (of $\delta$), but we have already seen in previous sections that the outcome depends on (a)~the number of the remaining rounds in the rotational phase, i.e. $T-T_1$, and also on whether (b)~$T-T_1$ is odd or even. The adjustment towards the ideal weights is more pronounced as $T-T_1$ increases, but an odd number of remaining rounds puts more emphasis on one of the options, thus discouraging the adjustment to the ideal weights. If the rotational phase is not long enough, the optimal adjustment may be partial, and the rotating weights assume values that lie between the greedy rotational weights and the ideal ones, as observed in the examples.
\end{enumerate}

A somewhat singular situation that might raise concerns of not falling into the above typology of penalties is the following: What if the game in the first $T_1$ rounds is as in \ref{step2}, which calls for a possible adjustment in round~1, but it then approaches the ideal weights as in~\ref{lab:3}\ref{lab:3b}. If the assumptions of \ref{lab:3}\ref{lab:3b} are valid, the first weight in the intersection area, say $w'$, is smaller than $1/(1+\beta)$ (so as to be in the intersection area), and greater than $1/(1+\sqrt{\beta})$ (so as to better approach the ideal weights from below). The more $w'$ approaches $1/(1+\beta)$, the more unlikely is the adjustment $\delta$ of type~\ref{step2} to be non zero.

A final note on the optimization of loss functions, which result from the above penalty patterns: All these loss functions are sums of terms of the form $f(k)$ or $f(k-2 \delta)$. For example, penalties as in pattern No. 2, i.e. $\left( \ell^0_1, \ell^1_1, \ldots \ell^{T-1}_1 \right) = (1-\delta,1,\ldots,1)$, end up with a loss equal to $F_{T-1}(\delta)$, where function $F$ is given by (\ref{eq:fcapitalfunction}), and can be expressed in terms of the q-Gamma and q-digamma functions as in (\ref{eq:fcapitalfunction2}). Even so, these functions are no more than a sum of terms of  of the form $f(k)$ or $f(k-2 \delta)$, as already mentioned. Pattern 3a is slightly more complicated, but it essentially remains of the same form. 

To the best of our knowledge there is no closed form solution for the optimization of the objective functions that result from patterns 2 and 3a. On the other hand, they are single valued functions, which can easily be maximized by using any non-linear maximization approach, which clearly is out of the scope of this paper.

On the contrary, there is a closed form (albeit very complex) solution for Pattern 3b. Firstly, this pattern can be simplified by calculating the rotation phase entry weight that results after $T_1$ greedy rounds, and using it as a starting weight for the rest of the calculation, which aims at optimizing the rotation phase only. Furthermore,  due to the periodicity the resulting objective function contains only three $f$ terms, since all odd valued round losses are equal and all even valued round losses are also equal. If $T-T_1$ is odd, the objective function is
\[
g(\delta) \equiv w_1 \delta + (1-w_1) (1-\delta) + \frac{T-T_1}{2}\left( \frac{w_1}{w_1+(1-w_1) \beta^{1-2 \delta}}
+  \frac{w_1 \beta^{2 \delta}}{w_1 \beta^{2 \delta}+1-w_1} \right)
\]
where $w_1$ is the weight after the first $T_1$ greedy rounds. There is a similar expression for even $T-T_1$. It is possible to maximize $g(\delta)$ by solving $g'(\delta)=0$. Since $g'(\delta)$ is a function of $\beta^{2\delta}$ only, we can set $x=\beta^{2\delta}$ and then solve the equation for $x$. The solution consists of four roots (with very long expressions) w.r.t. $x$, and thus a closed form solution can be found.

\section{Conclusions}
We have presented various aspects of the solution to the problem of finding  the  worst performance of the Hedge algorithm. We have shown that despite initial appearances, which point towards NP-completeness, the optimal strategy for the adversary remains almost greedy and binary. However, optimality often requires additional small non greedy adjustments, which depend on both the initial conditions and the long run of the game.

In short, the optimal adversarial strategy against Hedge consists of two discrete phases, if the game is long enough (so as to be able to accommodate both phases), and if the initial weight distribution is not uniform. These phases are: (a)~An almost greedy strategy that penalizes the maximum weight in each round, and ends up with an almost uniform weight distribution, and (b) a rotational behavior that penalizes each option periodically. 

The non greedy adjustments aim (i)~at properly entering the rotational phase and (ii)~at optimizing the performance of each cycle during rotation. The adjustments depend on the initial weight distribution and on the length of the game. A longer game is more likely to favor an adjustment. Interestingly, they also depend on the parity of the game length. However, adjustments are isolated abrupt (i.e. not gradual) events, as they take place in very few specific rounds of the game (if at all), i.e. at the beginning of the game or at the beginning of the second phase.

Our analysis also shows that from  a practical viewpoint the adversary can simply use the greedy strategy (of always maximally penalizing the currently maximum weight), and we provide margins of the adversary's performance loss w.r.t. the optimal strategy. However, in very long games the sub-optimality costs accumulate in each round.

\bibliographystyle{alpha}
\bibliography{refshedge}

\appendix
\section{Appendix}
\label{appendix1}

\begin{figure}
	\centering
  \includegraphics[width=.5\textwidth]{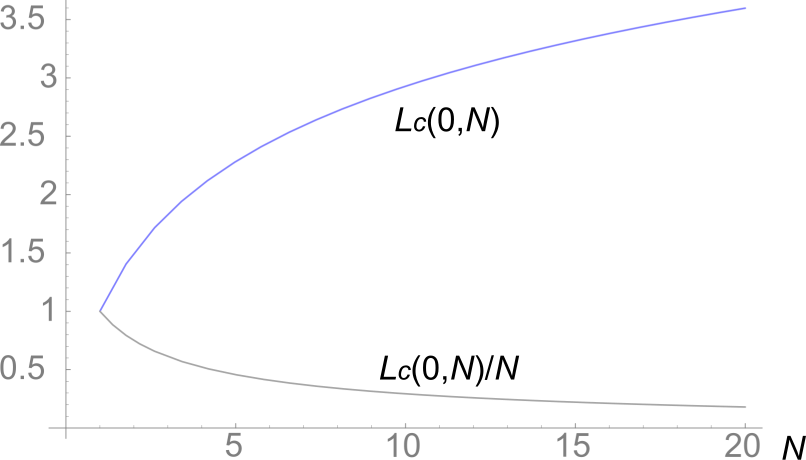}
  \caption{Per cycle maximum loss $L_c(0,N)$ and per step max.  $L_c(0,N) / N$ loss vs. $N$.}
  \label{fig:lcbound}
\end{figure}

\begin{example} \label{example:Noptionsexample}
Assume that the adversary uses   an alternating loss scheme as follows
\begin{eqnarray*}
\ell^0 &=& (1,0,0, \ldots, 0) \\
\ell^1 &=& (0,1,0, \ldots, 0) \\
\ell^2 &=& (0,0,1, \ldots, 0) \\
&& \ldots \\
\ell^{N-1} &=& (0,0,0, \ldots, 1) \\
\ell^{N} &=& (1,0,0, \ldots, 0) \\
\ell^{N+1} &=& (0,1,0, \ldots, 0) \\
&& \ldots
\end{eqnarray*}
The subsequent behavior of   Hedge   is shown on Table~\ref{table:alternatingN}. In fact any permutation of the first $N$ vectors will produce the same result, but without loss of generality we consider only the above scheme. The last column shows the single non-zero element of the loss vector in each round. Apparently the scheme repeats itself with a period (or cycle) of $N$ rounds. The total loss per cycle (consisting of $N$ rounds) is constant and equal to
\begin{table}
\centering
\noindent \begin{tabular}{lcccc}
\hline \hline
$t$ & $p_1^t$ & $p_2^t$ 
& $\ldots$ & $p_N^t$ \\
\hline \hline
$0$ & $w_1$ & $w_2$ 
& $\ldots$ & $w_N$  \\
$1$ & $\frac{w_1 \beta}{w_1 \beta + \sum_{i=2}^N w_i}$  & $\frac{w_2}{w_1 \beta + \sum_{i=2}^N w_i}$ 
& $\ldots$ & $\frac{w_N}{w_1 \beta + \sum_{i=2}^N w_i}$  \\
$2$ & $\frac{w_1 \beta}{\beta  \sum_{i=1}^2 w_i + \sum_{i=3}^N w_i}$  & $\frac{w_2  \beta}{\beta  \sum_{i=1}^2 w_i + \sum_{i=3}^N w_i}$ 
& $\ldots$ & $\frac{w_N}{\beta  \sum_{i=1}^2 w_i + \sum_{i=3}^N w_i}$  \\
$3$ & $\frac{w_1 \beta}{\beta  \sum_{i=1}^3 w_i + \sum_{i=4}^N w_i}$  & $\frac{w_2  \beta}{\beta  \sum_{i=1}^3 w_i + \sum_{i=4}^N w_i}$ 
& $\ldots$ & $\frac{w_N}{\beta  \sum_{i=1}^3 w_i + \sum_{i=4}^N w_i}$  \\
$\vdots$ &$\vdots$ &$\vdots$ 
&$\vdots$ &$\vdots$  \\
$N-1$ & $\frac{w_1 \beta}{\beta  \sum_{i=1}^{N-1} w_i + w_N}$  & $\frac{w_2  \beta}{\beta  \sum_{i=1}^{N-1} w_i + w_N}$ 
& $\ldots$ & $\frac{w_N}{\beta  \sum_{i=1}^{N-1} w_i + w_N}$  \\
$N$ & $\frac{w_1 \beta}{\beta  \sum_{i=1}^N w_i} = w_1$  & $\frac{w_2  \beta}{\beta  \sum_{i=1}^N w_i} = w_2$ 
& $\ldots$ & $\frac{w_N \beta}{\beta  \sum_{i=1}^N w_i} = w_N$  \\
\hline \hline
\end{tabular}
\caption{Hedge behaviour in the first $N+1$ rounds for a rotating loss scheme.}
\label{table:alternatingN}
\end{table}
\begin{eqnarray*}
&& L_c (N, \beta, w_1, \ldots, w_N)  \\
&=& w_1 + \frac{w_2}{w_1 \beta + \sum_{i=2}^N w_i} +  \frac{w_3}{\beta  \sum_{i=1}^2 w_i + \sum_{i=3}^N w_i} + \ldots + \frac{w_N}{\beta  \sum_{i=1}^{N-1} w_i + w_N}
\end{eqnarray*}
If $T=kN$ (where $k$ is an integer), the total Hedge loss is
\[
L_{H(\beta)} = \frac{T}{N} L_c (N, \beta, w_1, \ldots, w_N)
\]
and a simple modified expression can be written for any other integer $T$. 

On the other hand the following inequality holds for the cumulative loss per cycle 
\begin{equation}
1 \leq L_c (N, \beta, w_1, \ldots, w_N) \leq \frac{1}{\beta}
\end{equation}
which can be proved as follows: First, note that 
\[
\frac{w_j^0}{\beta  \sum_{i=1}^{j-1} w_i + \sum_{i=j}^{N} w_i} \geq w_j^0 \quad (j=2,3, \ldots, N)
\]
therefore 
\[ 
 L_c  (N, \beta, w_1, \ldots, w_N) \geq \sum_{j=1}^N w^0_j =1
\]
Regarding the upper bound
\[
\frac{w_j^0}{\beta  \sum_{i=1}^{j-1} w_i + \sum_{i=j}^{N} w_i} \leq \frac{w_j^0}{\beta  \sum_{i=1}^{j-1} w_i +\beta  \sum_{i=j}^{N} w_i} = \frac{1}{\beta} w_j^0 \quad (j=2,3, \ldots, N)
\]
therefore
\[
L_c  (N, \beta, w_1, \ldots, w_N) \leq \frac{1}{\beta} \sum_{j=1}^N w^0_j = \frac{1}{\beta}
\]
This means that if $T=kN$ the total loss is bounded as follows:
\begin{equation}
\frac{T}{N} \leq L_{H(\beta)} \leq \frac{T}{\beta N}
\end{equation}

In particular, for equal initial weights $w_1 = w_2 = \ldots = w_N = 1/N$ the cycle loss is simplified to
\begin{equation}
L_{ce}(N,\beta) \equiv L_c  (N, \beta, 1/N, \ldots, 1/N)=\sum_{i=0}^{N-1} \frac{1}{i \beta +N-i}
\label{eq:percyclelossforequalweights1}
\end{equation}
Sample values of the per round average loss $L_{ce}(N,\beta)/N$ obtained by using this formula are shown in Fig.~\ref{fig:lcfig}. Apparently, the per round loss decreases with the number of options, therefore if the adversary were allowed to choose the number of options, her choice would be $N=2$.
For $N=2$ options  the cycle loss reduces to
\[
L_{ce}(2,\beta) = \frac{1}{2} + \frac{1}{1 + \beta}
\]
while for arbitrary $N$
it can be shown to be equal to
\begin{equation}
L_{ce}(N,\beta) = - \frac{\psi(\frac{N}{-1+ \beta})- \psi(N+ \frac{N}{-1+ \beta})}{-1+ \beta}
\label{eq:percyclelossforequalweights}
\end{equation}
\begin{figure}
	\centering
  \includegraphics[width=.5\textwidth]{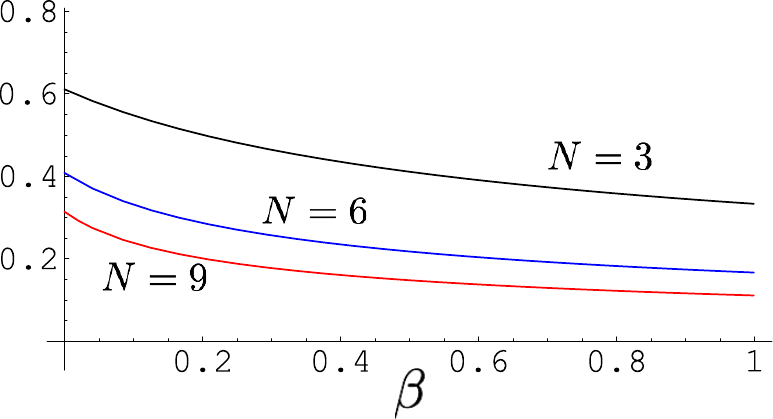}
  \caption{Per round loss $L_{ce}(N,\beta)/N$ vs. $\beta$ calculated by (\ref{eq:percyclelossforequalweights1}) for $N=3,6,9$.}
  \label{fig:lcfig}
\end{figure}
where $\psi(z) = \frac{\Gamma'(z)}{\Gamma(z)}$ is the digamma function. Sample values of $L_c$ are shown on Fig.~\ref{fig:lcfig}. For $\beta = 0$ the cycle loss is apparently maximized, and
\[
L_{ce}(N, 0) = \sum_{i=0}^{N-1} \frac{1}{i \beta +N-i} = \gamma + \psi(1+N)
\]
where $\gamma$ is the Euler-Mascheroni constant ($\gamma \approx 0.577216$). Note that the loss per round is thus upper bounded by $L_{ce}(N, 0) /N$, which is a decreasing function of $N$. Fig.~\ref{fig:lcbound} shows both $L_{ce}(N, 0)$ and $L_{ce}(N, 0) /N$.~$\Box$
\end{example}

\end{document}